\documentclass{article}

\usepackage{microtype}
\usepackage{graphicx}
\usepackage{subfigure}
\usepackage{booktabs} 

\usepackage{hyperref}

\usepackage[accepted]{icml2023}

\usepackage{amsmath}
\usepackage{amssymb}
\usepackage{mathtools}
\usepackage{amsthm}

\usepackage[capitalize,noabbrev]{cleveref}

\theoremstyle{plain}

\theoremstyle{definition}

\theoremstyle{remark}

\usepackage[textsize=tiny]{todonotes}

\icmltitlerunning{Demonstration-free Autonomous Reinforcement Learning via Implicit and Bidirectional Curriculum}

\usepackage{makecell}

\usepackage{bbm}

\usepackage{pifont}
\newcommand{\xmark}{\ding{55}}%

\colorlet{red}{black}
\colorlet{blue}{black}
\colorlet{magenta}{black}

\definecolor{cgreen}{RGB}{0, 255, 0}
\definecolor{xred}{RGB}{255, 0, 0}

\begin{document}

\twocolumn[
\icmltitle{Demonstration-free Autonomous Reinforcement Learning via \\ Implicit and Bidirectional Curriculum}



\icmlsetsymbol{equal}{*}

\begin{icmlauthorlist}
\icmlauthor{Jigang Kim}{equal,snu,aiis}
\icmlauthor{Daesol Cho}{equal,snu,aiis}
\icmlauthor{H. Jin Kim}{snu,asri}
\end{icmlauthorlist}

\icmlaffiliation{snu}{Seoul National University}
\icmlaffiliation{aiis}{Artificial Intelligence Institute of Seoul National University (AIIS)}
\icmlaffiliation{asri}{Automation and Systems Research Institute (ASRI)}

\icmlcorrespondingauthor{Jigang Kim}{jgkim2020@snu.ac.kr}
\icmlcorrespondingauthor{Daesol Cho}{dscho1234@snu.ac.kr}
\icmlcorrespondingauthor{H. Jin Kim}{hjinkim@snu.ac.kr}

\icmlkeywords{Machine Learning, ICML}

\vskip 0.3in
]



\printAffiliationsAndNotice{\icmlEqualContribution, \textcolor{magenta}{order decided by a coin toss.}} 

\begin{abstract}
While reinforcement learning (RL) has \textcolor{blue}{achieved} great success in acquiring complex skills solely from \textcolor{red}{environmental} interactions, it assumes that resets \textcolor{red}{to the initial state} are readily available at the end of each episode. \textcolor{red}{Such an assumption hinders the} autonomous learning of embodied agents \textcolor{red}{due to the time-consuming and cumbersome workarounds for resetting in the physical world}. Hence, there has been a growing interest in autonomous RL (ARL) methods that are capable of learning from non-episodic interactions. \textcolor{blue}{However,} existing works on ARL are limited by their reliance on prior data and are unable to learn in environments where task-relevant interactions are sparse. In contrast, we propose a demonstration-free ARL algorithm via \textbf{I}mplicit and \textbf{B}i-directional \textbf{C}urriculum \textcolor{red}{(\textbf{IBC})}. With an auxiliary agent that is conditionally activated upon learning progress and a bidirectional goal curriculum based on optimal transport, \textcolor{blue}{our method outperforms previous methods, even the ones that leverage demonstrations.}
\end{abstract}

\section{Introduction}\label{introduction}

Reinforcement learning (RL) has enabled interactive agents to learn complex skills in various domains with little to no prior knowledge \cite{andrychowicz2020learning, baker2019emergent, vinyals2019grandmaster, degrave2022magnetic}. However, existing algorithms assume an episodic setting where each trial begins from a state sampled from some fixed initial state \textcolor{red}{distribution, and they} are not designed to learn autonomously in the real world which involves continual, uninterrupted interaction. The challenge of applying RL in the real world often arises in robotics where the practitioner has to bridge the gap between the tools available (episodic RL) and the non-episodic nature of real-world learning. In most cases, a multitude of time-consuming and costly external interventions such as human supervision, task-specific scripted policies, and custom experimental setups are deployed to reset the environment after each trial \cite{kumar2016optimal, ha2020learning, nagabandi2020deep}. These challenges should be addressed from the algorithmic level by developing RL agents that can learn autonomously with minimal interventions.

Previous works on RL agents in the real world primarily involve a mechanism to handle resets and may leverage prior data along with additional consideration for reward assignment. \textcolor{red}{Reset mechanisms that prevent interventions by requesting a reset when necessary \cite{eysenbach2017leave, kim2022automating} are only viable if manual resets are readily available.} Under the non-episodic autonomous RL (ARL) framework \cite{sharma2021autonomous}, however, manual resets are not available on-demand \textcolor{red}{and the agent must learn} from continual interactions with no interventions. To overcome the challenge of the non-episodic setting, many previous methods rely on some form of prior data with varying degrees of privilege, ranging from the expert or sub-optimal trajectories \cite{sharma2022state, chen2022you} to examples of states of interest \cite{zhu2020ingredients}. However, a \textcolor{blue}{\emph{truly}} autonomous agent should be able to learn from scratch without \textcolor{blue}{external interventions and prior data.} To that end, we propose an ARL algorithm that can train a goal-conditioned RL policy without demonstrations under the non-episodic, sparse reward setting.

It has been well established that \textcolor{red}{existing RL algorithms do not perform well in the non-episodic setting} \cite{co2020ecological} since the agent is unable to repeatedly practice for the evaluation task. A common framework for extending conventional RL to the non-episodic setting is to alternate between multiple objectives within one continual interaction, effectively dividing it into multiple episodes. Typically, the forward episode attempts the original objective and the backward episode follows an auxiliary objective that provides an anchor for the forward episode with a good initialization. An obvious choice for the auxiliary objective is to return to the initial state distribution \cite{eysenbach2017leave}. However, this is not always \textcolor{blue}{the} optimal choice as it wastes valuable transitions on \textcolor{red}{returning} all the way back to the initial state. Instead, it can be set to match other distributions such as the states observed in expert demonstrations, \cite{sharma2022state} or maximize the state diversity \cite{zhu2020ingredients} for better sample efficiency or robustness.

\begin{figure}[t]
\centerline{\includegraphics[width=\linewidth]{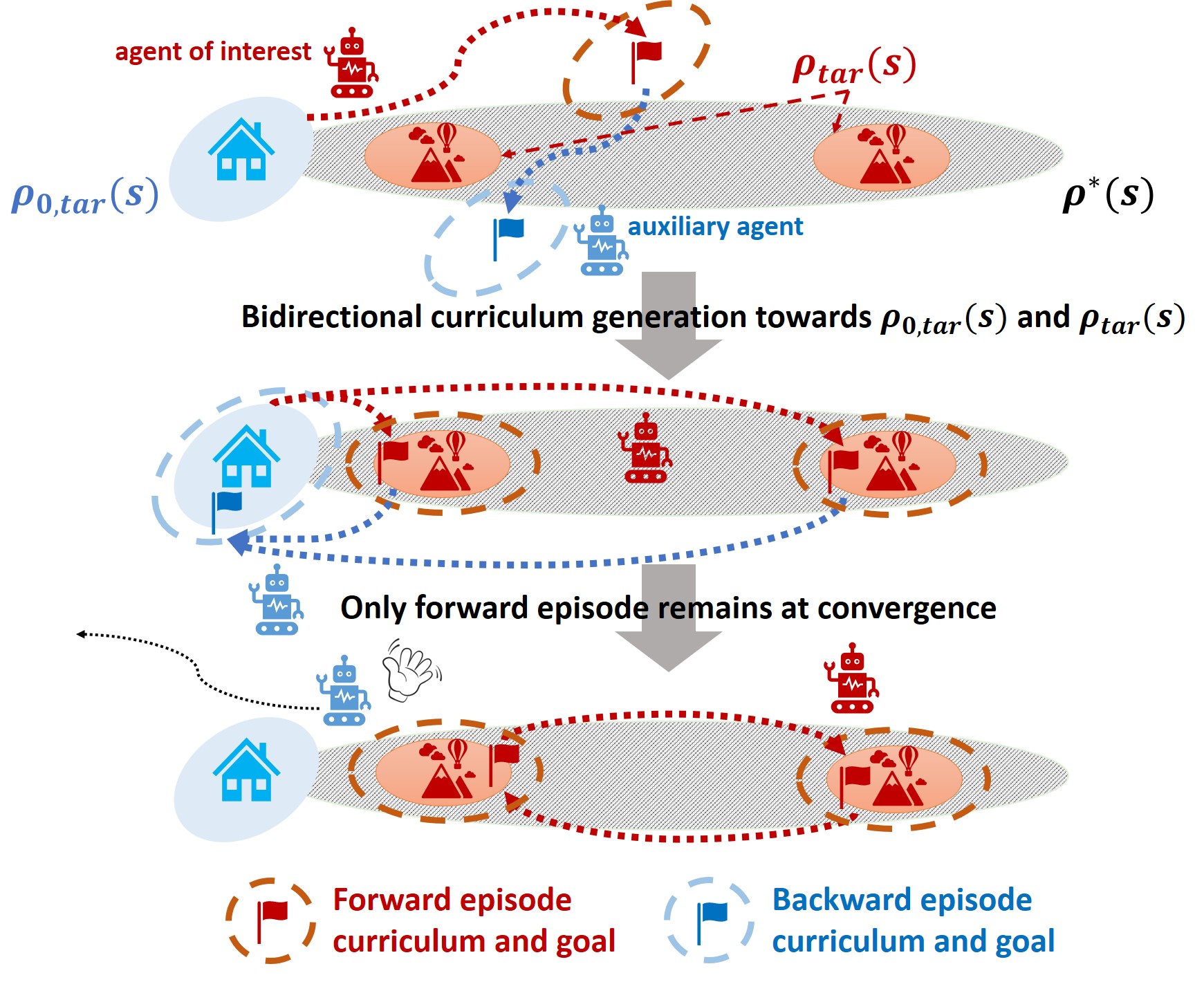}} 
\caption{IBC proposes a bidirectional curriculum for both forward and backward episodes. The auxiliary agent is no longer activated after the agent of interest becomes capable.} 
\label{fig:thumnail}
\end{figure}

We consider a \textcolor{blue}{conditionally activated} auxiliary agent that returns to the initial state based on our observation that providing a strong anchor is crucial, especially if the task of interest involves an interaction that is sparse and unlikely to occur by chance in the non-episodic setting. Under the proposed method, the agent of interest is initially dependent on the auxiliary agent but becomes less reliant on it as training progresses in an implicit curriculum. \textcolor{red}{When the agent of interest becomes capable, forward episodes can be rolled out consecutively without the auxiliary agent intervening} and more transitions are devoted to training the agent of interest leading to better sample efficiency. While the auxiliary agent initially provides a strong foundation, additional guidance is needed to successfully train the agent of interest. Since the agent of interest is goal-conditioned and must learn without prior data, we generate curriculum goals that do not \textcolor{blue}{rely on} demonstrations or predetermined \textcolor{blue}{curriculum. Specifically,} we propose a bidirectional goal curriculum scheme to simultaneously select appropriate goals for the forward (agent of interest) and backward (auxiliary agent) episodes. To do so, we employ a curriculum based on the optimal transport between the desired goals and the candidate states sampled from past trajectories in the replay buffer \textcolor{blue}{to} jointly optimize over the forward and backward curriculum goals.

The main contribution of our work is in proposing a demonstration-free ARL algorithm via \textbf{I}mplicit and \textbf{B}i-directional \textbf{C}urriculum (\textbf{IBC}). \textcolor{red}{Evaluations} in established ARL benchmarks and \textcolor{blue}{in} RL environments \textcolor{blue}{modified} for the ARL setting \textcolor{red}{show that} our method outperforms existing methods. Further \textcolor{blue}{analyses} and ablation studies reveal that the proposed implicit curriculum (auxiliary agent) and explicit curriculum (bidirectional goal curriculum) are well-formed and necessary to successfully learn in the demonstration-free, non-episodic setting. To summarize, the key takeaways from our work are as follows:
\begin{itemize}
    \item \textcolor{red}{To the best of our knowledge, IBC is the first algorithm for non-episodic RL that can consistently learn without manual resets and demonstrations by leveraging curriculum learning.}
    \item \textcolor{red}{We propose a conditionally activated} auxiliary agent and a bidirectional goal curriculum based on optimal transport to guide the agent of interest.
    \item \textcolor{red}{In} various environments, IBC achieves state-of-the-art performance \textcolor{red}{against previous} methods, including \textcolor{red}{even the} ones that leverage prior data.
\end{itemize}

\section{Related Works}\label{related_works}

Autonomy in RL has gained much interest as RL is increasingly applied to various real-world robotics applications. Many practical \textcolor{red}{applications adopted} task-specific workarounds to implement resets in the real world with varying levels of automation -- from human supervision and custom experiment setups \cite{yahya2017collective, zeng2020tossingbot} to scripted actions and pre-trained networks \cite{sharma2020emergent, thananjeyan2021recovery}. Since then, \textcolor{red}{several} algorithms for reset-free RL have been proposed, drawing inspiration from various topics in RL such as multi-task RL \cite{gupta2021reset, walkedon}, multi-stage RL \cite{smith2020avid, xu2022dexterous}, curriculum learning \cite{sharma2021vaprl}, and unsupervised skill discovery \cite{xu2020continual}. Recently, the autonomous RL (ARL) framework formally defined the non-episodic RL \textcolor{red}{setting. Several} works have sought to address some variation of \textcolor{red}{the ARL} problem such as leveraging prior data or human preference to enable single-life and lifelong learning \cite{chen2022you, lu2020reset} or \textcolor{blue}{to handle} irreversibility in the environment \cite{xie2022ask}. \textcolor{blue}{However,} we notice a lack of \textcolor{blue}{\emph{truly}} autonomous agents in existing methods and propose an ARL algorithm for demonstration-free, non-episodic RL in ergodic environments.

\textcolor{red}{A common framework for replacing manual resets is to alternate} between forward and backward episodes \cite{han2015learning}. Not all such methods are capable of non-episodic RL as \textcolor{red}{some of them} are not entirely reset-free and instead focus on reducing manual resets through backward episodes \cite{eysenbach2017leave,kim2022automating}. Reset-free methods that learn a separate policy to reset to diverse initial states \cite{zhu2020ingredients,xu2020continual} are only viable in environments where task-relevant interactions are likely to occur by chance and either require prior data such as states of interest or are geared towards acquiring behavior primitives for downstream tasks. MEDAL \cite{sharma2022state} and VaPRL \cite{sharma2021vaprl} are directly comparable to our \textcolor{red}{method, but} MEDAL returns to the state distribution of the optimal policy instead of the initial state which requires expert demonstrations. While VaPRL can technically operate without demonstrations, we have found that removing them results in significant performance degradation in practice. This is likely due to the subgoal curricula scheme \textcolor{red}{proposed} in VaPRL which relies on demonstrations both for gathering good goal candidates and in calculating the cost for the goal selection process.

Curriculum learning has been deployed in RL to improve sample efficiency, encourage exploration, and solve complex multi-stage tasks \cite{narvekar2020curriculum}. Such strengths are also desirable in the non-episodic setting. Curriculum in \textcolor{red}{episodic} RL often involves distribution matching to some desired task distribution \cite{ren2019exploration, klink2022curriculum, huang2022curriculum, cho2023outcome} and task difficulty \cite{florensa2018automatic, sukhbaatar2017intrinsic, portelas2020teacher, jiang2021prioritized}. \textcolor{red}{However, these methods are not designed for the non-episodic setting. To address this, we propose an auxiliary agent and bidirectional goal curriculum to incorporate both task difficulty and task distribution matching. The auxiliary agent gradually fades away in an implicit curriculum conditioned on the learning progress (success rate) of the agent of interest. To apply goal curriculum in the non-episodic setting, we generate curriculum goals not only for the task goal (forward episode) but also for the initial state (backward episode) based on the Wasserstein distance metric.}

\section{Preliminary}\label{preliminary}

\subsection{Autonomous Reinforcement Learning}

We assume an ergodic environment for the demonstration-free, non-episodic setting, similar to many previous \textcolor{blue}{works on} autonomous RL (ARL). We consider the Markov decision process (MDP) \textcolor{magenta}{$\mathcal{M} = (\mathcal{S,G,A,P}, r, \gamma, \rho_{0})$}, where $\mathcal{S}$ denotes the state space, $\mathcal{G}$ the goal space, $\mathcal{A}$ the action space, $\mathcal{P}(s'|s,a)$ the transition dynamics, $\gamma$ the discount factor, and $\rho_{0}$ the initial state distribution of the evaluation setting. The learning algorithm $\mathbb{A}$ is defined as $\mathbb{A}:\left\{s_j, a_j, r_j, s_{j+1}\right\}_{j=0}^t \mapsto\left\{a_t, \pi_t(\cdot|s)\right\}$, which maps the collected data until time $t$ to an action $a_t$ \textcolor{blue}{to be applied during the non-episodic training and its current best guess of the optimal evaluation policy $\pi_t(\cdot|s)$}.

Typical implementations of RL algorithms \textcolor{blue}{(episodic)} involve thousands or millions of sampling $s_0 \sim \rho_{0}(s)$, which require manual resets at the end of every \textcolor{blue}{episode}. However, under the ARL framework \textcolor{blue}{(non-episodic)}, the initial state $s_0 \sim \rho_0(s)$ is sampled only once at the beginning and the agent interacts with the environment through the actions \textcolor{blue}{$a_t$} determined by \textcolor{blue}{the} algorithm $\mathbb{A}$ until $t \xrightarrow{} \infty$. 

ARL defines \textcolor{blue}{the} \emph{Deployed Policy Evaluation metric}, which measures \textcolor{blue}{how fast the policy $\pi_t$ improves} in terms of the evaluation performance for a given task: 
\begin{equation}\label{eqn:arl}    
    \mathbb{D}(\mathbb{A})=\sum_{t=0}^{\infty} \Bigl[J\left(\pi^*\right)-J\left(\pi_t\right)\Bigr]
\end{equation}
where $J(\pi)= \mathbb{E}_{\rho_0, \pi, \mathcal{P}}\left[\sum_{t=0}^\infty \gamma^t r\left(s_t, a_t\right)\right]$, and $\pi^*$ is the optimal policy. The goal of algorithm $\mathbb{A}$ is to \textcolor{red}{minimize $\mathbb{D}(\mathbb{A})$} by learning as fast as possible.

\subsection{Surrogate Objective for Curriculum-based RL}

We replace the original RL objective with \textcolor{blue}{a} surrogate objective to be utilized for curriculum \textcolor{blue}{generation} in Section \ref{method} and describe it in detail. Let $\mathcal{T}$ be the joint distribution of some initial state $s_0$ and goal $g$. Then, the \textcolor{blue}{original objective} $\max_\pi J(\pi)$ can be represented as,
\begin{equation}\label{eqn:value_ftn}
    \max_\pi V^\pi\left(\mathcal{T}\right):=\underset{\left(s_0, g\right) \sim \mathcal{T}}{\mathbb{E}}\left[V^\pi\left(s_0, g\right)\right]
\end{equation}
where $V^\pi\left(s_0, g\right)$ is the goal-\textcolor{blue}{conditioned} value function.

Our approach relies on the following generalizability condition \cite{florensa2018automatic, luo2018algorithmic, asadi2018lipschitz, ren2019exploration} that is characterized by the Lipschitz continuity-based \textcolor{blue}{assumption}: 
\begin{equation}\label{eqn:lipschitz}
    \left|V^\pi\left(\mathcal{T}^{\prime}\right)- V^\pi(\mathcal{T})\right| \leq L \cdot D\left(\mathcal{T}, \mathcal{T}^{\prime}\right)        
\end{equation}
where \textcolor{blue}{$L$ is the Lipschitz constant and $D(\mathcal{T},\mathcal{T'})=\inf_{\mu\in\Gamma(\mathcal{T}, \mathcal{T'})}(\mathbb{E}_\mu[d((s_0, g),(s_0', g'))])$ is the Wasserstein distance based on the distance metric $d(\cdot, \cdot)$. $\Gamma\left(\mathcal{T}, \mathcal{T'}\right)$ denotes the set of all possible transport plans $\mu$.}

Under Eq (\ref{eqn:lipschitz}), optimizing Eq (\ref{eqn:value_ftn}) can be relaxed into the following lower-bound maximization,
\begin{equation}\label{eqn:surrogate}
    \max _{\mathcal{T}, \pi} \textcolor{red}{\Bigl[V^\pi(\mathcal{T})-L \cdot D\left(\mathcal{T}, \mathcal{T}^*\right)\Bigr]}
\end{equation}
where $(s_0^*, g^*) \sim \mathcal{T}^*$ is the joint distribution of the target initial state $s_0^*$ and target goal state $g^*$. \textcolor{magenta}{Intuitively, it maximizes the policy performance and closeness to $\mathcal{T}^*$, which results in a task curriculum with increasing difficulty.}

\section{Method}\label{method}

For \textcolor{blue}{a \emph{truly}} autonomous RL without external interventions and human supervision, we introduce 1) \textcolor{blue}{a conditionally activated} auxiliary agent ($\pi_a$) that aids the forward agent ($\pi_f$) and 2) a bidirectional curriculum generation process for both forward and auxiliary agents \textcolor{blue}{that enables} non-episodic RL without demonstrations via calibrated guidance.

\subsection{Non-Episodic RL with an Auxiliary Agent}\label{method:non_episodic_RL}

\begin{figure}[t]
    \centering
    \includegraphics[width=0.9\columnwidth]{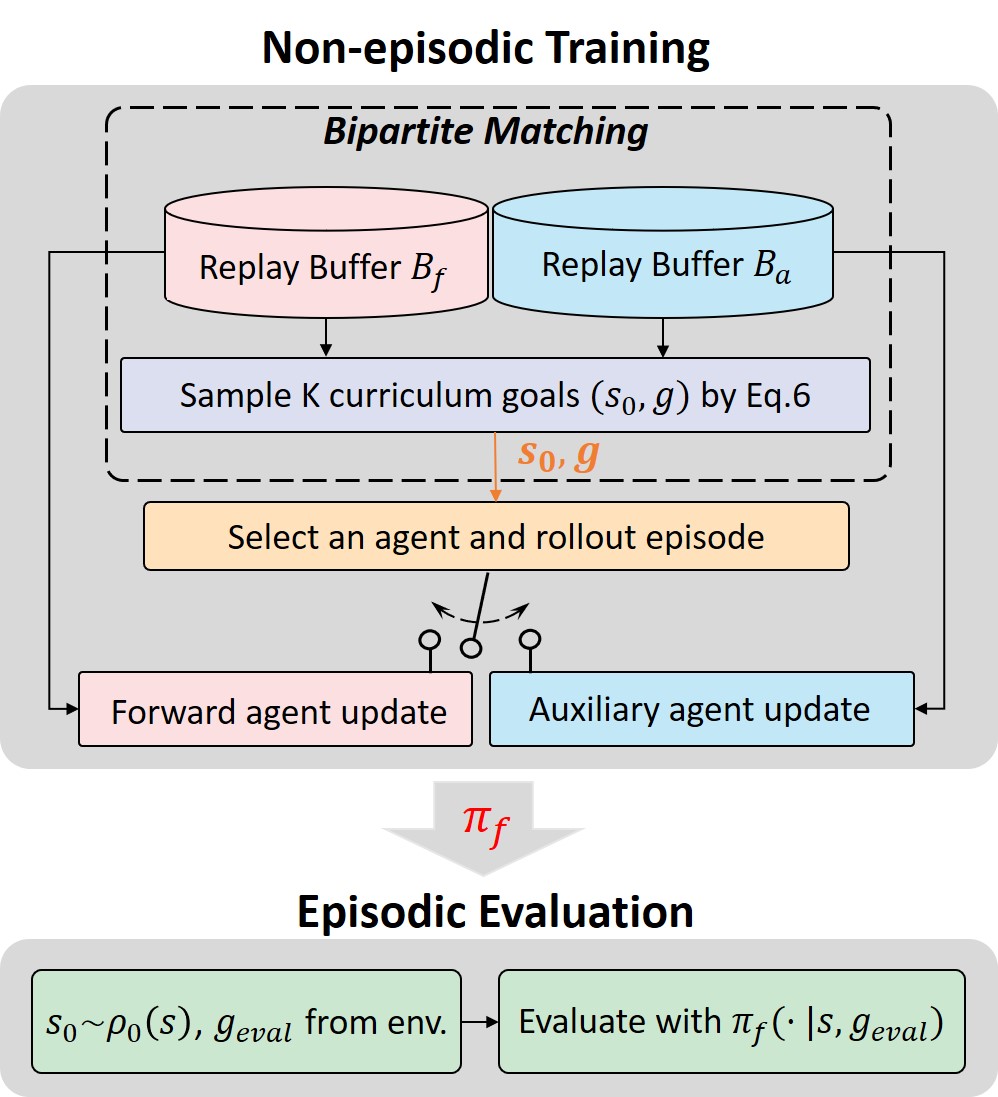}  
  \caption{Overview of the proposed method, IBC.}
  \label{fig:method_diagram}
\end{figure}

\textcolor{magenta}{
During non-episodic training, we alternate between the two agents such that the auxiliary agent guides the forward agent only when necessary. Specifically, we conditionally activate the auxiliary agent when the forward agent has failed at the given goal state such that the auxiliary agent gradually disappears as the forward agent improves which results in better sample efficiency. Let us consider the hypothetical setting where the forward agent is fully capable and the auxiliary agent does not intervene at all. Under this setting, the forward agent repeatedly attempts its target goal states $s_{g^*} \sim \rho_{tar}(s)$ without resets. Thus, the agent is no longer restricted by $\rho_0(s)$ unlike in episodic settings and we can consider a better initial state distribution by appropriately designing $\rho_{tar}(s)$.
}

\textcolor{magenta}{
Interestingly, a previous work \cite{kakade2002approximately} provides theoretical grounds that $\rho_0(s)$ close to $\rho^*(s)$ enables efficient training in RL, where $\rho^*(s)$ denotes the state marginal distribution of the optimal policy $\pi^*$. If we set $\rho_{tar}(s)$ to be a subset of $\rho^*(s)$ from the optimal policy that achieves the evaluation goal $g_{eval}$, we can approximately satisfy this ideal initial state distribution. Note that the target goal $s_{g^*}$ achieved by the forward agent policy $\pi_f$ from the previous rollout becomes the initial state for the next rollout.
}

\textcolor{magenta}{
In practice, it suffices for $\rho_{tar}(s)$, which is only used for bidirectional curriculum and not for RL, to contain a minimal number of key points that roughly outline the task to be adequate for the goal curriculum generation. This is because the curriculum goals effectively “fill in the blanks” by proposing past states from the replay buffer that are close to $\rho_{tar}(s)$. Typically, specifying $\rho_{tar}(s)$ requires only a handful of samples ($\sim 10$) from $\rho_0(s)$ and $g_{eval}$ combined to approximate $\rho^*(s)$. For some tasks, it suffices to specify $\rho_{tar}(s)$ with a single example from $\rho_0(s)$ and $g_{eval}$ each. Unlike previous ARL methods, we do not require demonstrations with thousands of transitions or access to the expert policy.
}

\begin{figure}[t]
\centerline{\includegraphics[width=\linewidth]{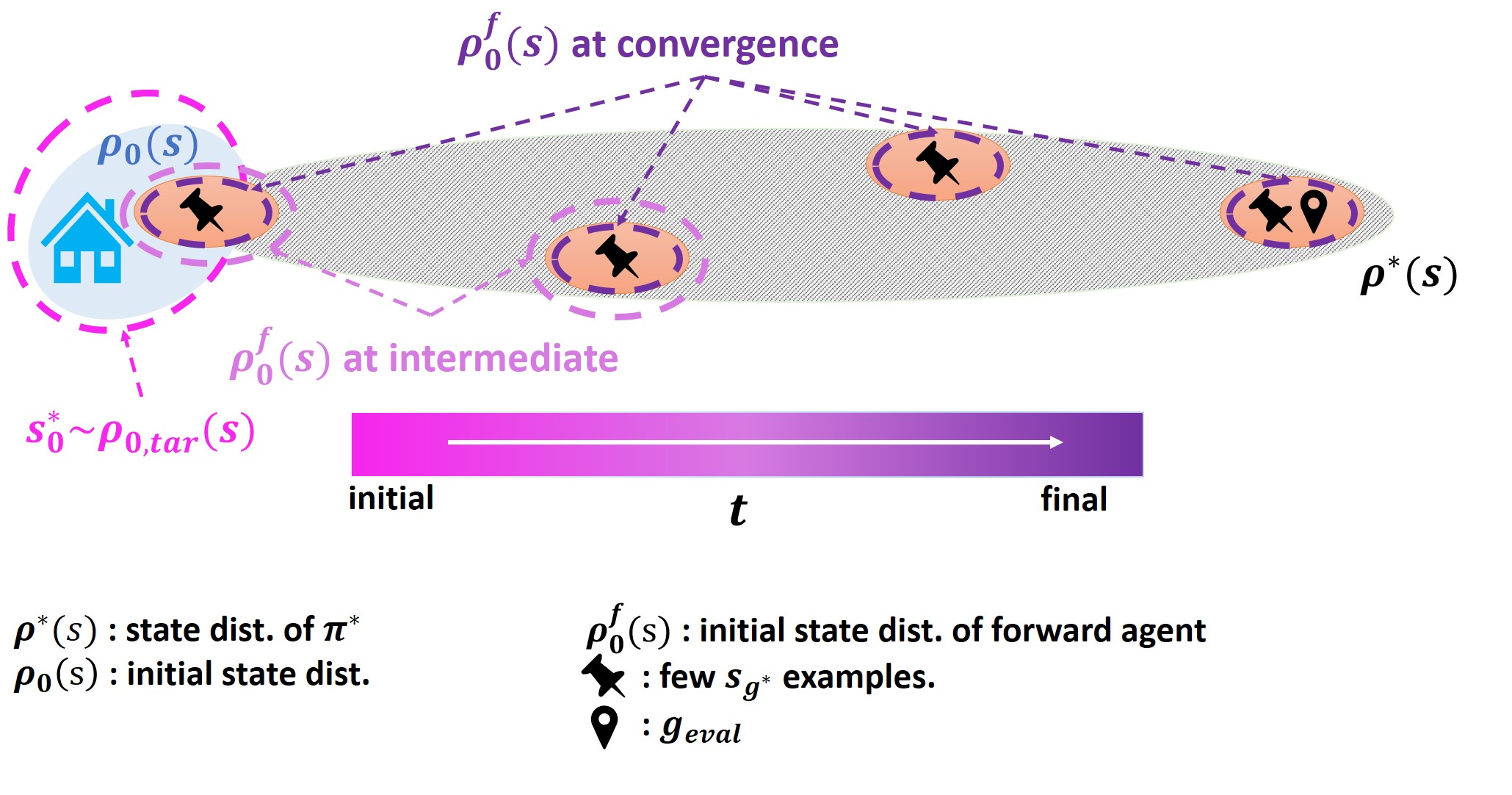}} 
\caption{Visualization of $\rho_0^f(s)$ at various timesteps. As training progresses, \textcolor{blue}{the initial state distribution of the forward agent $\rho_0^f(s)$} gradually shifts from $\rho_0(s)$ to \textcolor{magenta}{$\rho_{tar}(s)$}.} 
\vskip -0.3cm
\label{fig:vis-forward-initial}
\end{figure}

\textcolor{magenta}{
Until now, we have considered the setting where $\pi_f$ has converged and is fully capable. However, most of the rollouts by $\pi_f$ before convergence will lead the agent to an arbitrary state rather than $s_{g^*}$, leading to highly-varying initial states for the next rollout which results in unstable learning. For this reason, we need an auxiliary agent that provides an anchor and guides the forward agent. More precisely, the auxiliary agent tries to bring the forward agent back to the set of target initial states $s_0^* \sim \rho_{0,tar}(s)$. Even though $\rho_{0,tar}(s)$ can be an arbitrary set of states that are useful for the repeated practice of the forward agent, we set $\rho_{0,tar}(s)$ to include the environmental initial state distribution $\rho_0(s)$. This is because providing a strong anchor is crucial in practice and the evaluation will be performed from $\rho_0(s)$.
}

\textcolor{magenta}{
While the proposed auxiliary agent resembles the standard backward agent from previous literature, there are two key differences. First, the auxiliary agent is activated only when the forward agent fails at the given curriculum goal such that the initial state for the forward agent $\rho_0^f(s)$ gradually evolves from $\rho_0(s)$ to $\rho_{0,tar}(s)$ in an implicit curriculum as $\pi_f$ improves (Figure \ref{fig:vis-forward-initial}). This is by design such that the forward agent approximately satisfies the theoretically-grounded ideal initial state condition at convergence. Second, the auxiliary agent does not directly return to $\rho_{0,tar}(s)$ which encompasses $\rho_0(s)$, but to the intermediate goal for $s_0^*$ obtained from the bidirectional goal curriculum. Thus, the goals proposed to the auxiliary agent are more diverse than the typical backward agent.
}

\begin{algorithm}[h]
   \caption{IBC}
   \label{alg:overview}
\begin{algorithmic}[1]
    \STATE {\bfseries Input:} $\widehat{\mathcal{T}}$, $\widehat{\mathcal{T}}^*$, $\mathcal{B}_f$, $\pi_f$, $Q^{\pi_f}$, $\mathcal{B}_a$, $\pi_a$, $Q^{\pi_a}$
    
    \STATE $s \sim \rho_0(s)$ by $\mathtt{env.reset()}$, set $\mathcal{O}$ to $\{f\}$

    $\mathcal{O}$ denotes agent selection ($\{f\}$:forward, $\{a\}$:auxiliary)

    \WHILE{not done}
    \STATE \textcolor{blue}{get curriculum goal $g_{\mathcal{O}} \sim \widehat{\mathcal{T}}$}
    \WHILE{\textcolor{blue}{until (reach $g_{\mathcal{O}}$ or max episode steps)}}
    \STATE $a \leftarrow \pi_{\mathcal{O}}(\cdot \vert s, g_{\mathcal{O}})$
    \STATE $s' \leftarrow \mathcal{P}(s'|s,a), r \leftarrow r(s,a,g_{\mathcal{O}})$
    \STATE $\mathcal{B}_{\mathcal{O}} \leftarrow \mathcal{B}_{\mathcal{O}} \cup (s,g_{\mathcal{O}},a,r,s')$
    \STATE update $\pi_{\mathcal{O}}, Q^{\pi_{\mathcal{O}}}$
    \STATE $s \leftarrow s'$
    \ENDWHILE
    
    \FOR{once every $N$ \textcolor{blue}{iteration}}
        \STATE update $\widehat{\mathcal{T}}$ according to Eq (\ref{eqn:surrogate}) by solving Eq (\ref{eqn:hgg})
    \ENDFOR

    \STATE \textbf{if} $\mathcal{O}$ was $\{a\}$ \textbf{then} set $\mathcal{O}$ to $\{f\}$ \\
    \STATE \textbf{else} \textbf{if} $\pi_f$ succeeded \textbf{then} \textcolor{blue}{keep $\mathcal{O}$ as} $\{f\}$ \\
    \STATE \textbf{else} set $\mathcal{O}$ to $\{a\}$

    \ENDWHILE
   
\end{algorithmic}
\end{algorithm}

\subsection{Bidirectional Curriculum Generation}\label{method:curriculum}

\textcolor{blue}{While our non-episodic training process involving an auxiliary agent, \textcolor{magenta}{$\rho_{0,tar}(s)$}, and \textcolor{magenta}{$\rho_{tar}(s)$} approximately satisfies the ideal initial state condition,} it might not be sufficient for autonomous training in environments where target states are difficult to be \textcolor{red}{achieved} from scratch. \textcolor{blue}{Thus, we need to find intermediate goals that can guide the learning of the agent. To find such goals without relying on demonstrations, the candidates must be obtained from past trajectories with highly varying initial states due to non-episodic training. We propose a bidirectional goal curriculum based on the surrogate problem (Eq (\ref{eqn:surrogate})) for both forward and auxiliary agents without relying on demonstrations in the non-episodic setting.}

\textcolor{blue}{For autonomous curriculum generation,} we sample the candidates for $\mathcal{T}$ from past states in the replay buffer $\mathcal{B}$. To prevent a degenerate solution in the curriculum selection process, \textcolor{blue}{a diversity constraint} is incorporated such that for every trajectory $\tau = (s_0, ..., s_{t_{final}}) \in \mathcal{B}$, at most one state can be chosen for $\mathcal{T}$. Then, Eq (\ref{eqn:surrogate}) is transformed as follows,
\begin{equation}\label{eqn:with_constraint}
\begin{aligned} 
    \max_{\pi_f, \mathcal{T}} \:\: & \textcolor{red}{\Bigl[V^{\pi_f}(\mathcal{T})-L \cdot D\left(\mathcal{T}, \mathcal{T}^*\right)\Bigr]} \\
    \text { s.t. } & \sum_{t} \mathbbm{1}\left[\left(s_0, \phi_f\left(s_t\right)\right) \in \mathcal{T}\right] \leq 1, \quad s_0, s_t \in \tau, \forall \tau \in \mathcal{B} \\
\end{aligned}
\end{equation}
where $\phi(\cdot)$ is a mapping function that abstracts the state space into the goal space. To solve Eq (\ref{eqn:with_constraint}), we iteratively update $\mathcal{T}$ and policies $\pi_f, \pi_a$ until $\pi_f$ achieves a desirable evaluation performance. \textcolor{blue}{The policy optimization is simply achieved by applying off-the-shelf RL algorithms such as SAC \cite{haarnoja2018soft}. The optimization of $\mathcal{T}$ is defined by the Wasserstein Barycenter problem augmented with a value bias term.}

\textcolor{blue}{Inspired by \citet{ren2019exploration}, we enforce $\mathcal{T}$ and $\mathcal{T}^*$ to be a set of $K$ particles ($\vert\widehat{\mathcal{T}}\vert=\vert\widehat{\mathcal{T}}^*\vert=K$) where $(s_0, \textcolor{magenta}{g})^i \sim \widehat{\mathcal{T}}$, and $(s_0^*, \textcolor{magenta}{\phi(s_{g^*})})^i \sim \widehat{\mathcal{T}}^*$, rather than parameterizing their distribution. Then, to address the Wasserstein Barycenter problem (Eq (\ref{eqn:with_constraint})) in the combinatorial setting, we assign candidates for $\widehat{\mathcal{T}}$ to $\widehat{\mathcal{T}}^*$ via the following bipartite matching problem:}

\textcolor{magenta}{
\begin{equation}\label{eqn:hgg}
    \min_{\tau^i=\{s_t^i, \forall t \}\in\mathcal{B}} \sum_{\left(s_0^*, s_{g^*}\right)^i} w\left(\left(s_0^*, s_{g^*}\right)^i, \tau^i\right)    
\end{equation}
}
\textcolor{blue}{where $w(\cdot, \cdot)$ becomes}

\textcolor{magenta}{
\begin{equation}\label{eqn:hgg_cost}
\begin{split}
  &w\left(\left(s_0^*, s_{g^*}\right)^i, \tau^i\right):=c\left\|\phi_a(s_0^{*,i}) -\phi_a(s_0^i)\right\|_2 \\ 
  &+\min _t\left(\left\|\phi_f(s_{g^*}^i)-\phi_f(s_t^i)\right\|_2-\frac{1}{L} V^{\pi_f}(s_0^i, \phi_f(s_t^i))\right), \\
\end{split}
\end{equation}
}
\textcolor{blue}{when} we define the distance metric $d((s,g), (s',g'))$ \textcolor{blue}{from Eq (\ref{eqn:lipschitz})} as $c\left\|\phi_a(s)-\phi_a(s')\right\|_2 + \left\|g-g'\right\|_2$ ($c$ is a hyperparameter). \textcolor{blue}{With the costs $w$ defined according to Eq (\ref{eqn:hgg_cost}), we can construct a bipartite graph $\mathbf{G}(\{\mathbf{V}_a, \mathbf{V}_b\},\mathbf{E})$. Let $\mathbf{V}_a$ be the set of nodes representing candidates for $\widehat{\mathcal{T}}$ and $\mathbf{V}_b$ be the set of nodes for $\widehat{\mathcal{T}}^*$. The \textcolor{red}{weights} of the edges are defined as $\mathbf{E}(v_a,v_b) = -w(v_a,v_b)$, where $v_a \in \mathbf{V}_a$ and $v_b \in \mathbf{V}_b$.}

To solve the bipartite matching problem, the Minimum Cost Maximum Flow algorithm is utilized to find K edges with the minimum combined cost of connecting $\mathbf{V}_a$ and $\mathbf{V}_b$ \citep{ahuja1993}. \textcolor{blue}{The resulting $K$ forward curriculum goals will be} proposed towards a region of the state space considered to be close to \textcolor{magenta}{$s_{g^*} \sim \rho_{tar}(s)$} \textcolor{red}{and} within the capability of the forward agent \textcolor{blue}{as indicated by the value bias term}. Similarly, the \textcolor{blue}{$K$ auxiliary curriculum goals will be} proposed towards a region considered to be close to \textcolor{magenta}{$s_0^* \sim \rho_{0,tar}(s)$}.

\begin{table*}[t]
\caption{Conceptual comparison between our work and baseline algorithms.}
\label{table_comparison}
\centering
\resizebox{\textwidth}{!}{%
\begin{tabular}{c|cccc}
\noalign{\smallskip}\noalign{\smallskip}\Xhline{3\arrayrulewidth}
{} & Demo-free & Curriculum  & Agent Configuration  & Backward Towards \\
\Xhline{3\arrayrulewidth}
oracle RL & \textcolor{cgreen}{\checkmark} & \textcolor{xred}{\xmark} & single (SAC)  & N/A  \\
\hline
R3L & \textcolor{cgreen}{\checkmark} & \textcolor{xred}{\xmark}  &  forward (VICE, \citet{fu2018variational}) \& backward (RND, \citet{burda2018exploration}) & max$\mathcal{H}(s)$ for diverse states  \\
\hline
VaPRL & \textcolor{xred}{\xmark} & backward subgoal only (\textcolor{cgreen}{\checkmark}) &  single (SAC) & $\rho_0(s)$  \\
\hline
MEDAL & \textcolor{xred}{\xmark} & \textcolor{xred}{\xmark} &  forward (SAC) \& backward (VICE with $\rho^*(s)$) & $\rho^*(s)$ from expert demos  \\

\Xhline{3\arrayrulewidth}
\textbf{IBC(ours)} & \textcolor{cgreen}{\checkmark} & both forward \& backward (\textcolor{cgreen}{\checkmark}) & dual (forward \& auxiliary) $\xrightarrow{}$ single as training proceeds (SAC) &  \textcolor{magenta}{$\rho_{tar}(s)$} (a subset of $ \rho^*(s)$)  \\
\Xhline{3\arrayrulewidth}
\end{tabular}

}
\end{table*}

\begin{figure*}[t]
    \centering
    \subfigure[Fetch Pick\&Place\label{fig:eval-success-rate-pickandplace}]{\includegraphics[width=0.32\textwidth]{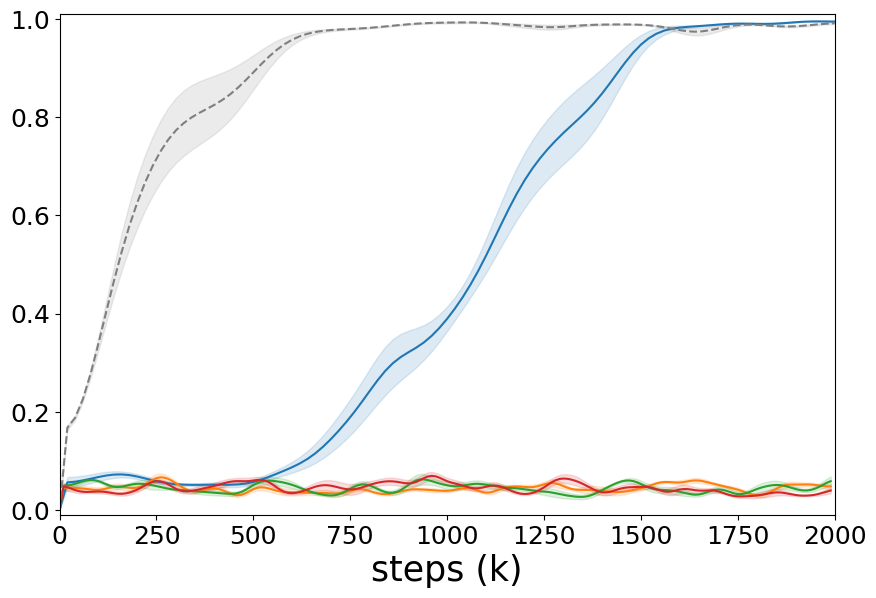}}
    \subfigure[Fetch Push\label{fig:eval-success-rate-push}]{\includegraphics[width=0.32\textwidth]{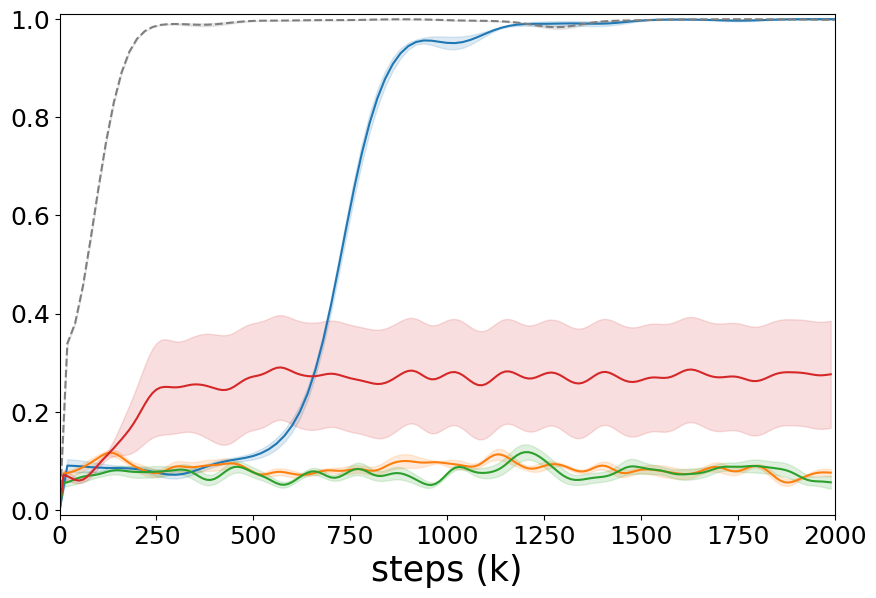}}
    \subfigure[Fetch Reach\label{fig:eval-success-rate-reach}]{\includegraphics[width=0.32\textwidth]{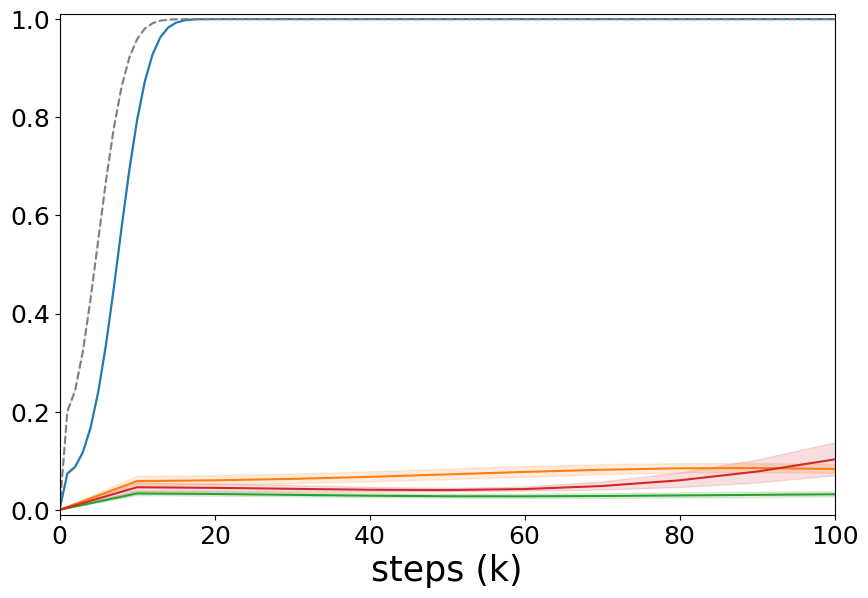}}
    \subfigure[Sawyer Door\label{fig:eval-success-rate-door}]{\includegraphics[width=0.32\textwidth]{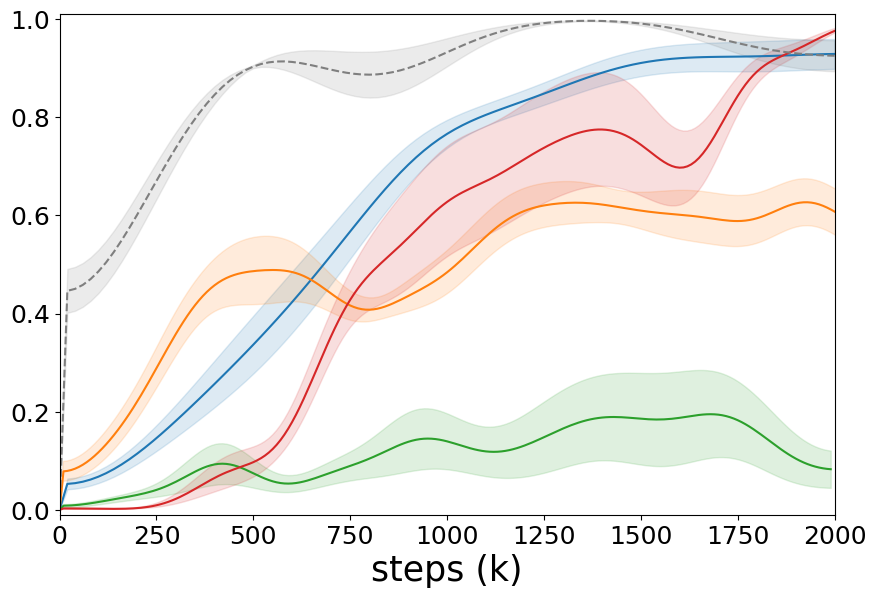}}
    \subfigure[Tabletop Manipulation\label{fig:eval-success-rate-table}]{\includegraphics[width=0.32\textwidth]{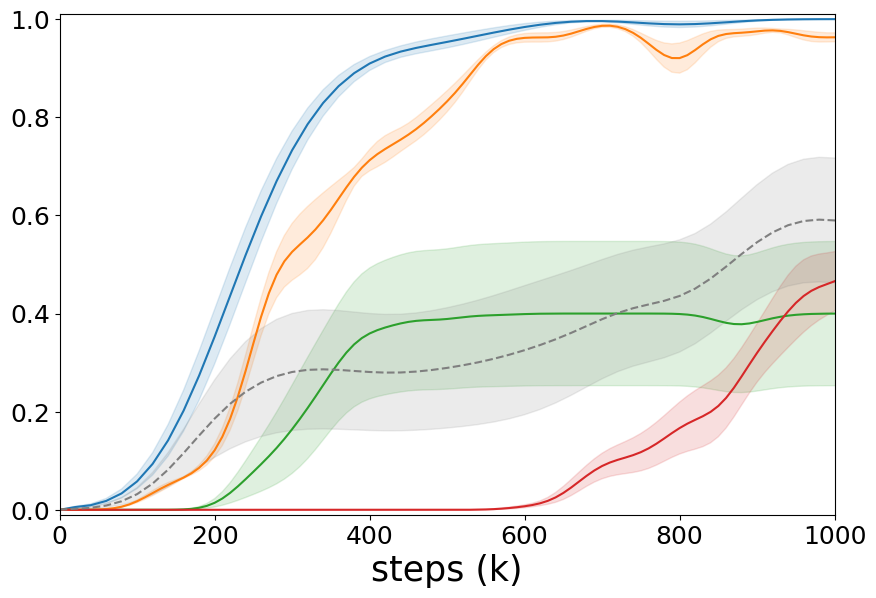}}
    \subfigure[Point-U-Maze\label{fig:eval-success-rate-umaze}]{\includegraphics[width=0.32\textwidth]{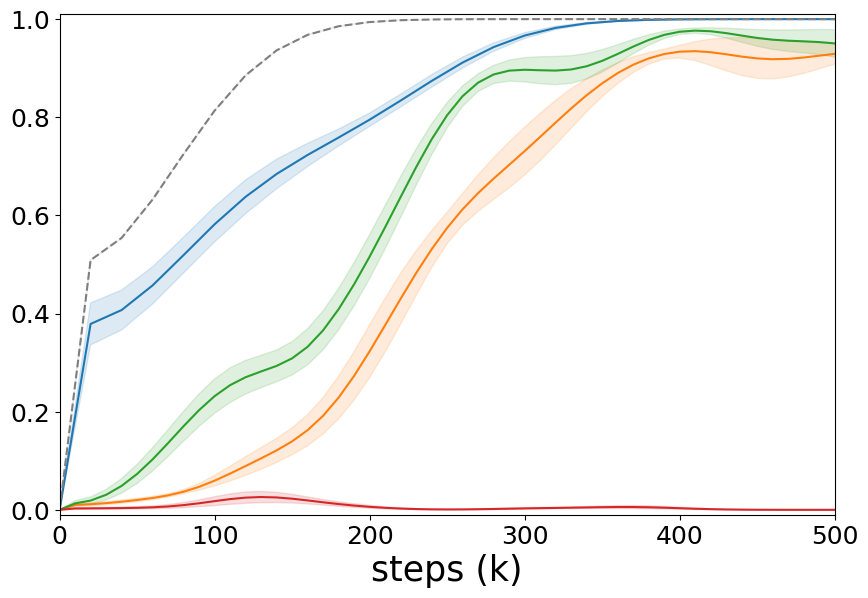}}
    
    \subfigure{\includegraphics[width=0.6\textwidth]{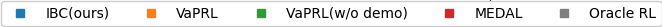}}
    
  \caption{Comparison of evaluation success rates of various algorithms. Shading indicates standard deviation across 5 seeds.}
  \label{fig:eval-success-rate}
\end{figure*}

\section{Experiment}\label{experiment}

We include six sparse reward environments to evaluate our method. Two environments -- Tabletop Manipulation, Sawyer Door -- are from established ARL benchmark, EARL \cite{sharma2021autonomous}, and the remaining four environments -- Fetch environments \cite{plappert2018multi}, Point-U-Maze -- are modified versions of existing MuJoCo-based OpenAI Gym environments \cite{todorov2012mujoco, brockman2016openai} for the ARL setting. These environments represent a mixture of robotic manipulation and locomotion tasks. Detailed descriptions of the environments are provided in Appendix \ref{appendix:experiment details}.

We compare with other previous methods designed for the ARL framework, which can be summarized as follows:

\textbf{MEDAL} \cite{sharma2022state} -- a backward agent that minimizes the distance between its state marginal distribution and the expert state distribution.

\textbf{VaPRL} \cite{sharma2021vaprl} -- value-based subgoal curricula towards the initial state distribution $\rho_0(s)$ \textcolor{blue}{during the backward episode}; amenable to demonstration-free setting, but reports on the version with demonstration data.

\textbf{oracle RL} -- a standard RL baseline such as SAC \cite{haarnoja2018soft} in an episodic setting with goal relabeling technique \cite{andrychowicz2017hindsight} common for sparse reward environments. 

There exist other ARL methods such as R3L \cite{zhu2020ingredients} but we did not include them as they are already outperformed by VaPRL and MEDAL. We summarize the conceptual comparison between our method and previous ARL baselines in Table 1.

\begin{figure*}[t]
    \centering
    \includegraphics[width=\textwidth]{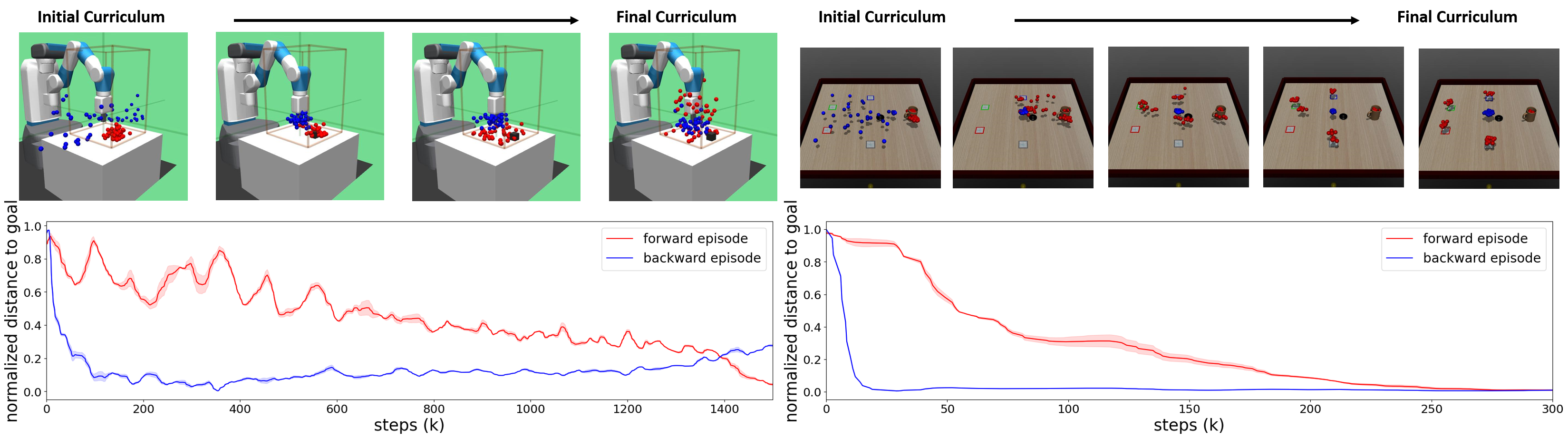}

  \caption{Visualization of the curriculum goals and their average normalized distance to assigned target goals (Left: Fetch Pick\&Place, Right: Tabletop Manipulation). The red and blue dots indicate the curriculum goals for the forward and auxiliary agents, respectively. Note that the exact positions of the robots and objects are meaningless; these are just rendered from their default states.}
  \label{fig:curriculum-visualize}
\end{figure*}

\subsection{Results and \textcolor{blue}{Analyses}}\label{experiment:evaluation-analysis}

We follow the evaluation setting similar to the EARL benchmark \cite{sharma2021autonomous}. Specifically, \textcolor{red}{the agent interacts} with the environment after \textcolor{blue}{initially being spawned at $s_0 \sim \rho_0(s)$ and occasionally being reset to $s_0 \sim \rho_0(s)$} after hundreds of thousands of steps. Since we focus on minimizing the deployed policy evaluation metric, $\mathbb{D}(\mathbb{A})$, we report on $J(\pi_t)$ in 10k \textcolor{blue}{training step} intervals by averaging returns from the policy over multiple evaluation episodes. \textcolor{magenta}{The code implementation of IBC and the instructions for reproducing the main result is available at \url{https://github.com/snu-larr/ibc_official}.}

\paragraph{Evaluation results.} As shown in Figure \ref{fig:eval-success-rate}, the proposed method achieves state-of-the-art performance against other baselines, without requiring any demonstration data and even achieving comparable average return (success rate) to the oracle RL (episodic RL setting). Although some prior works such as VaPRL and MEDAL utilize nearly expert-level demonstration data, they have difficulty in environments where the task-relevant interactions are very sparse in the non-episodic setting or \textcolor{blue}{the evaluation goals $g_{eval}$} are uniformly spread over some region rather than a few points such as Fetch environments. Furthermore, these methods are somewhat sensitive to the composition of the demonstration data in practice, \textcolor{blue}{which is detailed in} Appendix \ref{appendix:more_experiments}. For a fair comparison with our method, we also evaluated a version of VaPRL \textcolor{blue}{without demonstrations;} it performed noticeably worse than the original VaPRL.

\begin{figure}
    \centering
    \subfigure[Fetch Pick\&Place\label{fig:fb-ratio-pickandplace}]{\includegraphics[width=0.5\columnwidth]{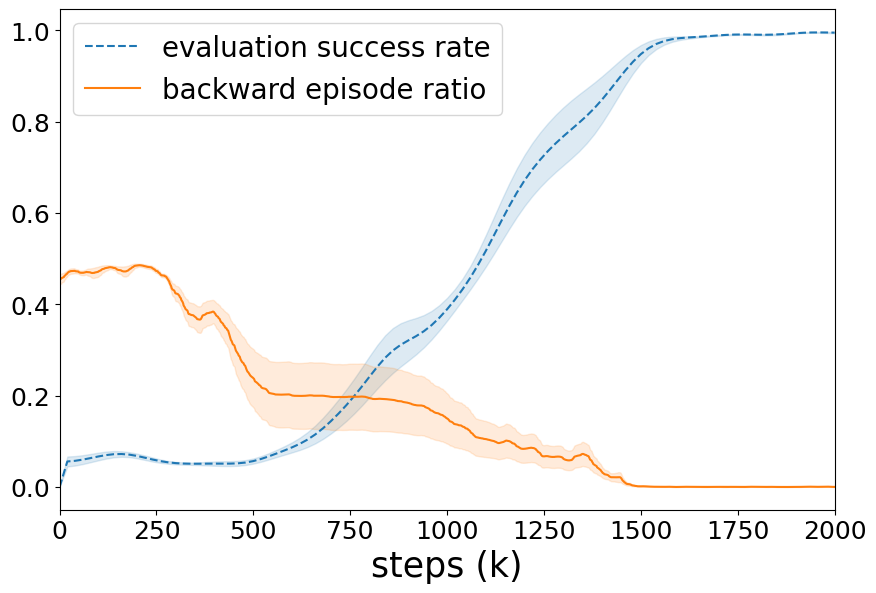}}
    \subfigure[Tabletop Manipulation\label{fig:fb-ratio-tabletop}]{\includegraphics[width=0.5\columnwidth]{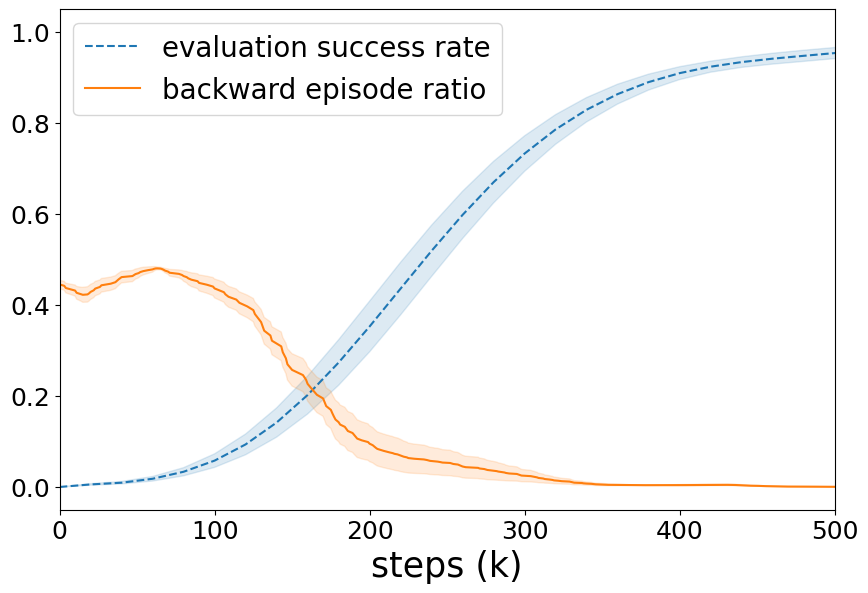}}

  \caption{Episode ratio of the auxiliary agent and evaluation success rate.}
  \label{fig:fb-ratio}
\end{figure}

To validate whether the \textcolor{blue}{intervention of the} auxiliary agent vanishes as training proceeds, we plot the episode ratio of the auxiliary agent \textcolor{blue}{within the latest 1k episodes. As shown in Figure \ref{fig:fb-ratio}, the auxiliary agent does not intervene when the forward agent is fully trained.}

\paragraph{Bidirectional curriculum.} To validate whether the bidirectional curriculum goals are properly interpolated and \textcolor{blue}{eventually converge} to the desired target distributions, we \textcolor{blue}{evaluate} the progress of the curriculum goals qualitatively and quantitatively. To do so, we visualize the forward and auxiliary curriculum goals and plot the corresponding normalized distance averaged over target goals assigned by bipartite matching (Section \ref{method:curriculum}).

The plots in Figure \ref{fig:curriculum-visualize} demonstrate that the average distance to goals consistently decreases as training proceeds, which indicates that the curriculum goals for both forward and auxiliary agents have properly converged to \textcolor{blue}{their respective} target states. The visualizations in Figure \ref{fig:curriculum-visualize} provide further validation. Specifically, the forward curriculum goals gradually converge toward the \textcolor{magenta}{$\rho_{tar}(s)$}, which encompasses a region in the air and on the table for the Fetch Pick \& Place, and five discrete points for the Tabletop Manipulation, respectively. The auxiliary curriculum goals also converge to the target goal states \textcolor{magenta}{$\rho_{0,tar}(s)$}, initially. However, there is a gradual shift of the auxiliary curriculum goals towards $\rho^*(s)$ after initial convergence which is reflected in the slight increase in average distance to goals for the backward episode (\textcolor{magenta}{$\rho_{0,tar}(s)$}), especially \textcolor{blue}{visible in} the Fetch Pick \& Place environment. This is because the candidates for the backward curriculum goals, which eventually become the initial states for the forward agent, are obtained from both \textcolor{magenta}{$\rho_{0,tar}(s)$} and \textcolor{magenta}{$\rho_{tar}(s) \subset \rho^*(s)$} when the forward agent remains at intermediate proficiency ($\sim$50\%) for prolonged timesteps during training such as in Fetch Pick \& Place, but less so in Tabletop Manipulation.

\begin{figure}[t]
    \centering
    \subfigure[Fetch Pick\&Place\label{fig:ablation-pickandplace}]{\includegraphics[width=0.49\columnwidth]{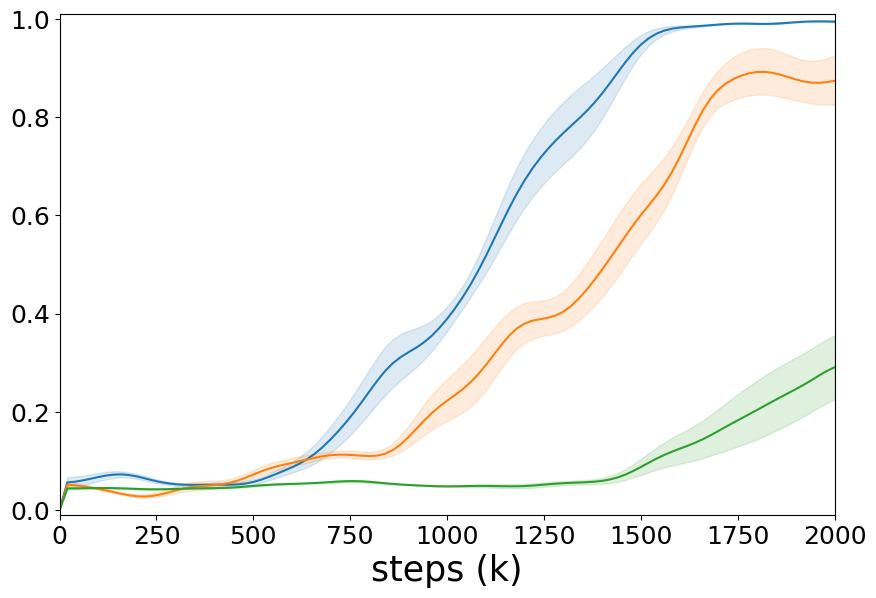}}    
    \subfigure[Sawyer Door\label{fig:ablation-door}]{\includegraphics[width=0.49\columnwidth]{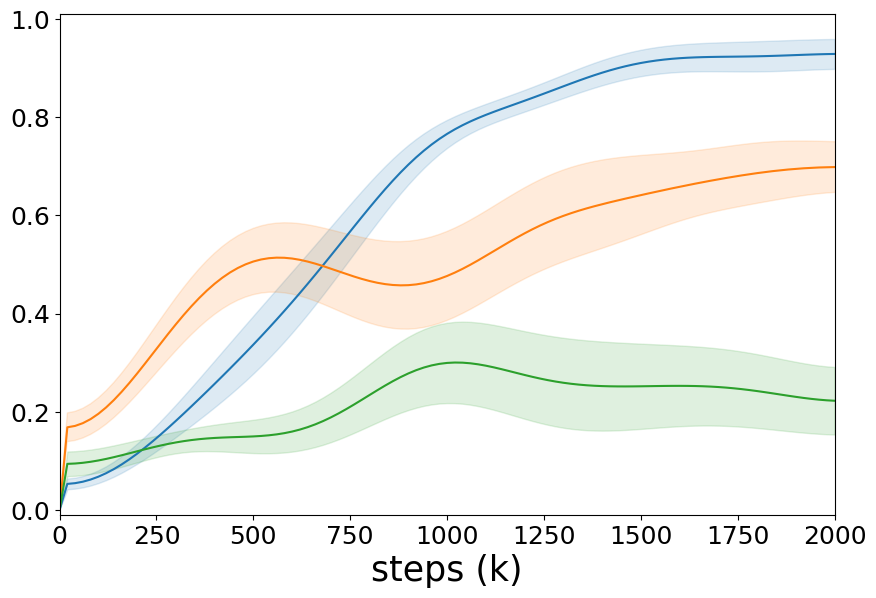}}
    \vskip -0.3cm
    \subfigure[Tabletop Manipulation\label{fig:ablation-table}]{\includegraphics[width=0.49\columnwidth]{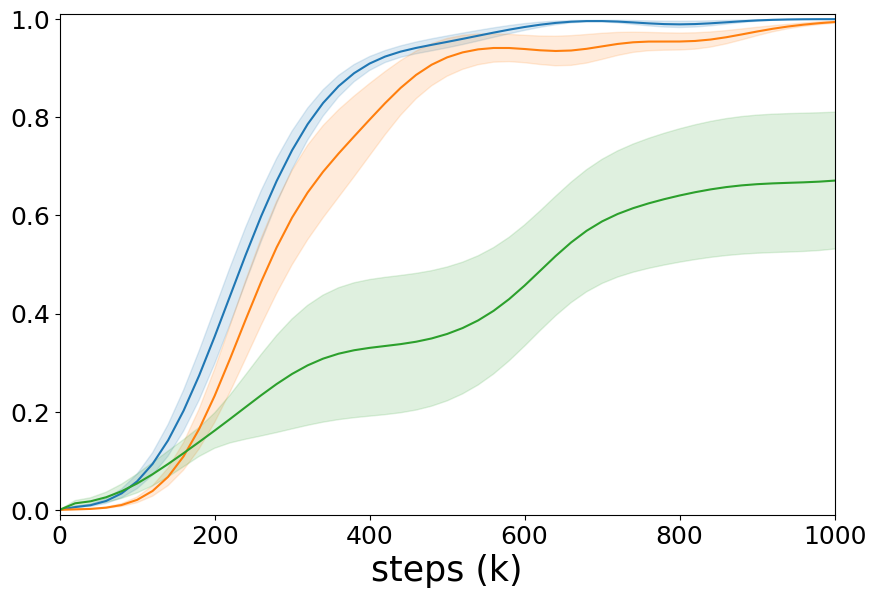}}
    \subfigure[Point-U-Maze\label{fig:ablation-umaze}]{\includegraphics[width=0.49\columnwidth]{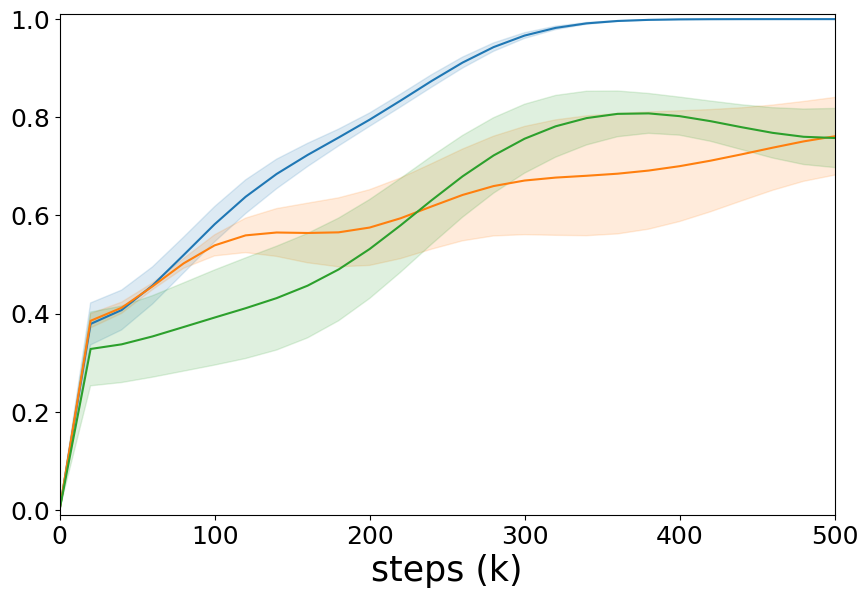}}
    \subfigure{\includegraphics[width=\columnwidth]{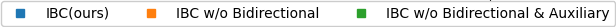}}
  \caption{Ablation study -- removing the bidirectional curriculum and auxiliary agent proposed in this work degrades performance.}
  \label{fig:ablation}
\end{figure}

\subsection{Ablation Study}\label{experiment:ablation}

To investigate the role of the \textcolor{blue}{goal} curriculum, we \textcolor{blue}{conduct} an experiment without the proposed bidirectional curriculum (\textbf{IBC w/o Bidirectional}). We additionally ablate the auxiliary agent to validate its effectiveness (\textbf{IBC w/o Bidirectional \& Auxiliary}). The latter corresponds to naive RL, where only a forward agent tries to optimize the task reward \textcolor{blue}{in a reset-free setting}, without any backward episodes to return to some region such as $\rho_0(s)$. 

As shown in Figure \ref{fig:ablation}, there is consistent performance degradation in most of the environments for \textbf{IBC w/o Bidirectional}, which demonstrates the importance of gradually guiding the forward agent from easier initial states and goals to difficult ones. For \textbf{IBC w/o Bidirectional \& Auxiliary}, there is an additional degradation in most of the environments, \textcolor{blue}{more so in object manipulation environments. The exact degree of the degradation may vary according to the given task and the choice of goal space mapping $\phi(s)$. In the case of Tabletop Manipulation, bidirectional curriculum generation does not have much effect on performance since the state space is relatively simple and does not benefit from intermediate curriculum goals. In the case of Point-U-Maze, the auxiliary agent is less effective since the initial state is quite far from the goal states.}

\subsection{Towards Reward-free Operation}\label{experiment:wihout_reward}

To further enhance the autonomy of our method, we investigate the viability of the reward-free setting. While the proposed method operates on sparse rewards under the goal-conditioned setting which \textcolor{red}{involves} minimal human effort (defining the threshold for success), eliminating the need for explicit reward specification can be desirable, especially when dealing with high-dimensional inputs such as images. Prior work based on control as inference framework for future event matching \cite{fu2018variational, eysenbach2021replacing} \textcolor{red}{enables} reward-free methods that can also be applied to goal-conditioned RL such as C-learning \cite{eysenbach2020c}. We evaluate the performance of a variant of our method that replaces the SAC agent with a C-learning agent. For the sake of simplicity, we do not implement bidirectional curriculum goals for this variant but note that it is trivial to do so.

We additionally report the normalized \textcolor{blue}{distance to goal} metric along with the success rate for this variant as it is based on future state matching and does not consider the threshold for success. Full results are available in Appendix \ref{appendix:more_experiments}. In most environments, the C-learning variant achieves high success rates and even matches the proposed method in some environments. \textcolor{blue}{However, there are environments with low success rates, albeit with visible improvements in terms of the normalized distance to goal metric}. This is because C-learning is reward-agnostic and tries to match the entirety of the desired future state, \textcolor{blue}{but some state elements may be} more important than others for success depending on the task. Instead of matching in the state space, modifying the reward-free method to operate in a goal space that autonomously evolves towards the state elements that are most salient for the given task can be a promising approach for future research.

\begin{figure}[t]
    \centering
    \subfigure[Fetch Pick\&Place\label{fig:reward-free-pickandplace}]{\includegraphics[width=0.5\columnwidth]{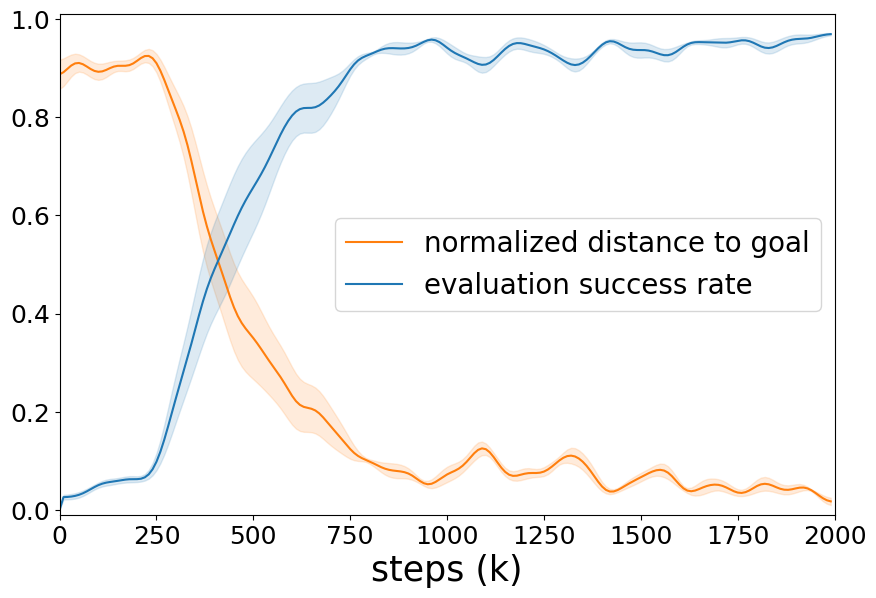}}
    \subfigure[Fetch Push\label{fig:reward-free-push}]{\includegraphics[width=0.5\columnwidth]{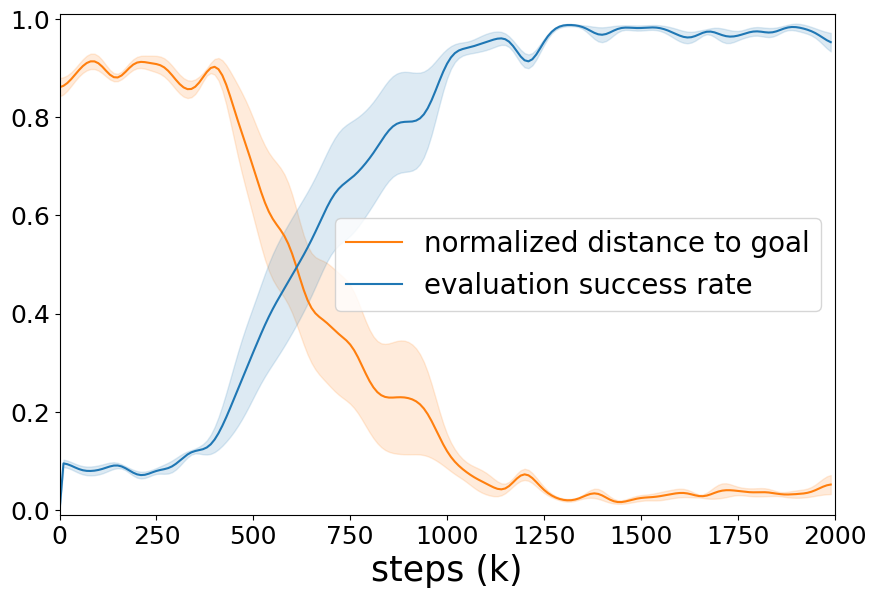}}
    \vskip -0.3cm
    \subfigure[Point-U-Maze\label{fig:reward-free-point}]{\includegraphics[width=0.5\columnwidth]{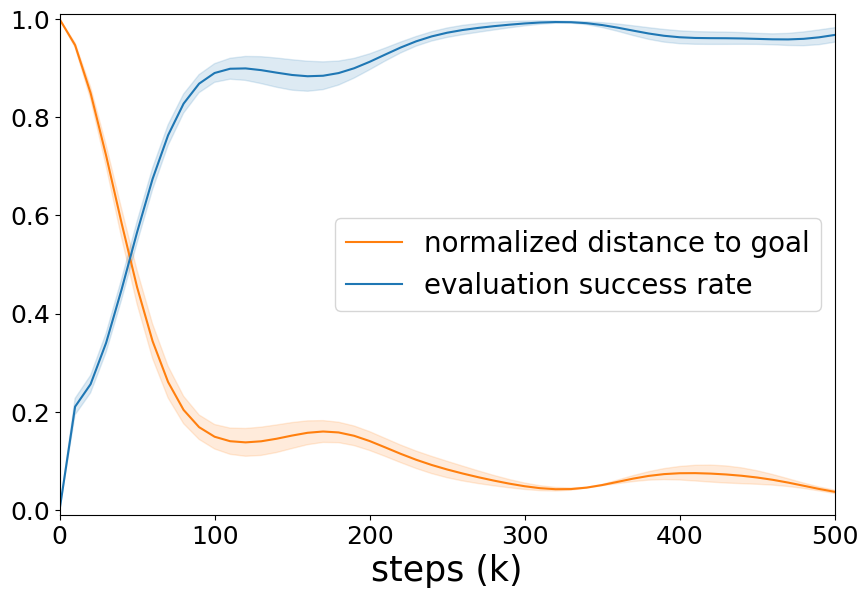}}
    \subfigure[Tabletop Manipulation\label{fig:reward-free-tabletop}]{\includegraphics[width=0.5\columnwidth]{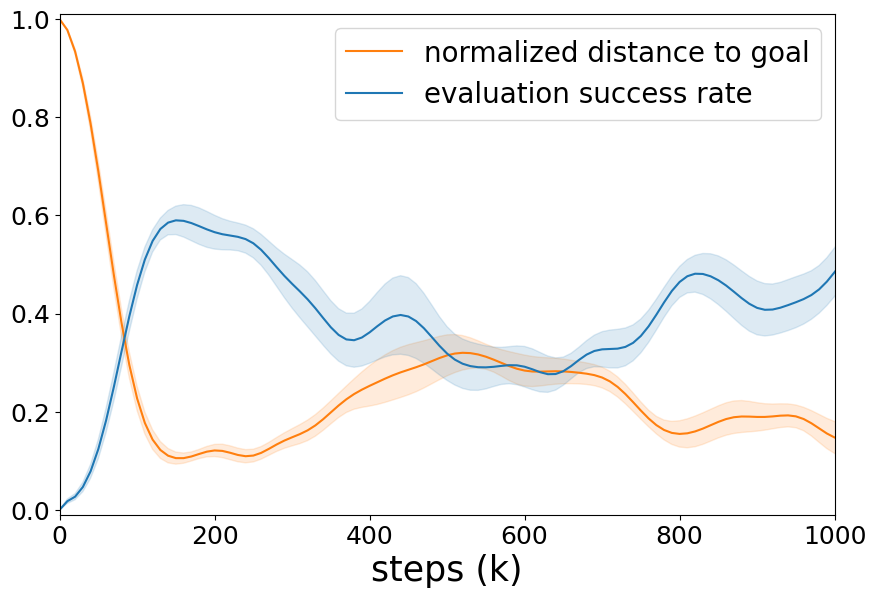}}
  \caption{Normalized distance to goal and evaluation success rate for the reward-free variant.}
  \label{fig:reward-free}
\end{figure}

\section{Conclusion}\label{conclusion}

In this work, we \textcolor{blue}{considered} a non-episodic RL setting where the agent should learn how to perform the given task autonomously without any external interventions such as manual resets and prior data. We proposed IBC, a demonstration-free autonomous learning algorithm based on implicit and bidirectional curriculum generation. We have shown that our method outperforms previous methods, both in terms of sample efficiency and final average success rate. Our method is limited to reversible environments and still requires minimal human inputs for specifying sparse rewards. We'd like to build upon our method towards the reward-free setting, which has shown some promise in our results, by adopting the contextual MDP framework and devising a task-relevant goal space curriculum \textcolor{blue}{discovery for} the reward-free setting.

\section{Acknowledgement}\label{acknowledgement}
\textcolor{magenta}{
This work was supported by the Korea Research Institute for Defense Technology Planning and Advancement (KRIT) Grant funded by the Defense Acquisition Program Administration(DAPA) (No. KRIT-CT-23-003, Development of AI tank crews based on deep reinforcement learning and establishment of virtual combat experiment)
}

\bibliography{main}
\bibliographystyle{icml2023}

\newpage
\appendix
\onecolumn


\section{Experimental Details}\label{appendix:experiment details}

\subsection{Environment}

\begin{itemize}
    \item \textbf{Sawyer Door}: We use the original EARL benchmark environment \cite{sharma2021autonomous}. The evaluation goal state \textcolor{blue}{$g_{eval}$} is the state where the door is fully closed and the target goal states \textcolor{magenta}{$s_{g^*} \sim \rho_{tar}(s)$} during the non-episodic training are set to the corresponding states for door hinge \textcolor{blue}{angles between -60 degrees (open) and 0 degrees (closed)}.
    \item \textbf{Tabletop Manipulation}: We use the original EARL benchmark environment \cite{sharma2021autonomous}. The evaluation goal states consist of 4 discrete points and the target goal states are 5 discrete points (4 discrete evaluation goal points \textcolor{blue}{+} 1 initial state of the object).
    \item \textbf{Point-U-Maze}: We used the 12 × 12 U-shaped maze environment where the initial position of the agent is $[0, 0]$ and the evaluation goal position is at the other end of the maze located at $[0, 8]$. The target goal states are randomly sampled in the feasible (not interfering with the maze walls, free space) state space. 
    \item \textbf{Fetch Pick\&Place, Fetch Push}: We modified the original Fetch environments from the gym-robotics package to convert it to a reversible (ergodic) setting by defining a constraint on the block position. The evaluation goals and target goal states are identical to the original Fetch Pick\&Place, Fetch Push environments.
    \item \textbf{Fetch Reach}: We use the original Fetch environment \textcolor{blue}{from the} gym-robotics package, where the evaluation goals and target goal states are obtained from \textcolor{blue}{uniformly sampled states} in the area encompassing a region in the air and on the table.
\end{itemize}

\begin{figure}[h]
    \centering
    \subfigure[Sawyer Door]{\includegraphics[width=0.16\columnwidth]{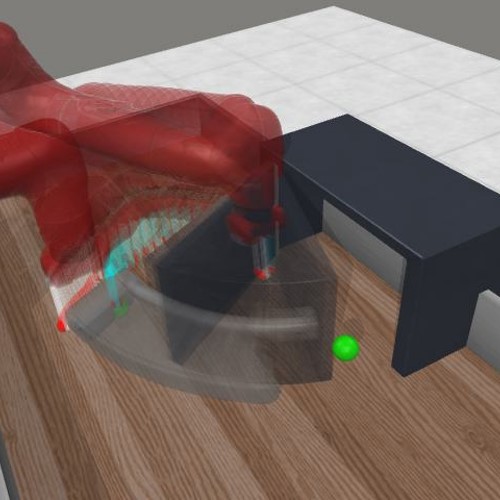}}
    \subfigure[Tabletop Manip.]{\includegraphics[width=0.16\columnwidth]{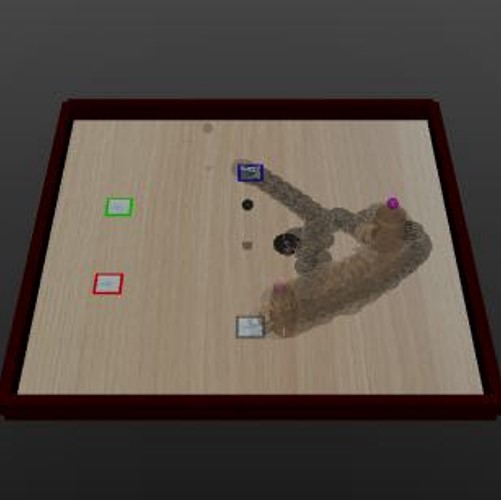}}
    \subfigure[Point-U-Maze]{\includegraphics[width=0.16\columnwidth]{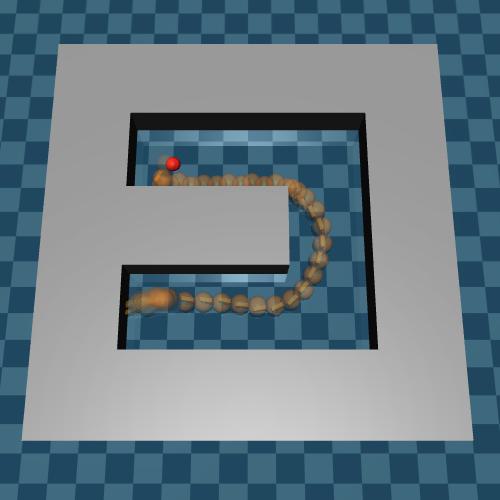}}
    \subfigure[Fetch Pick\&Place]{\includegraphics[width=0.16\columnwidth]{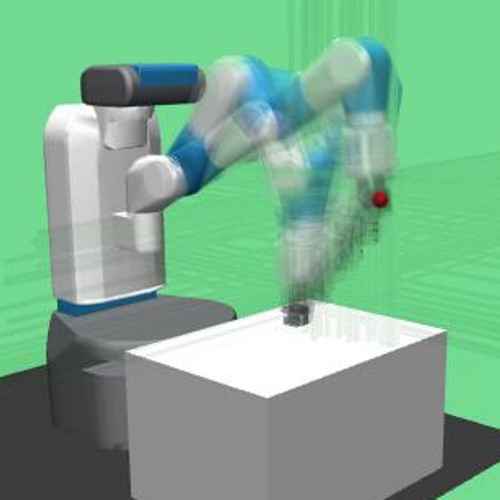}}
    \subfigure[Fetch Push]{\includegraphics[width=0.16\columnwidth]{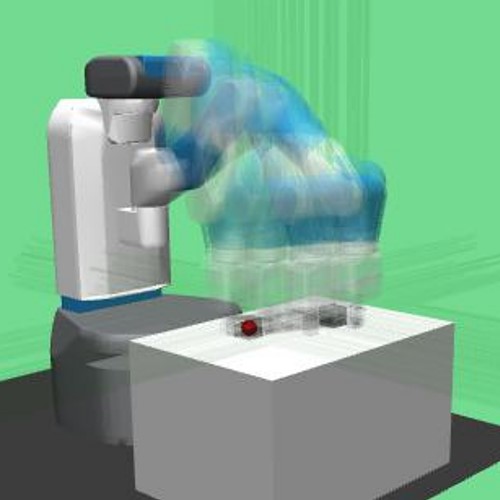}}
    \subfigure[Fetch Reach]{\includegraphics[width=0.16\columnwidth]{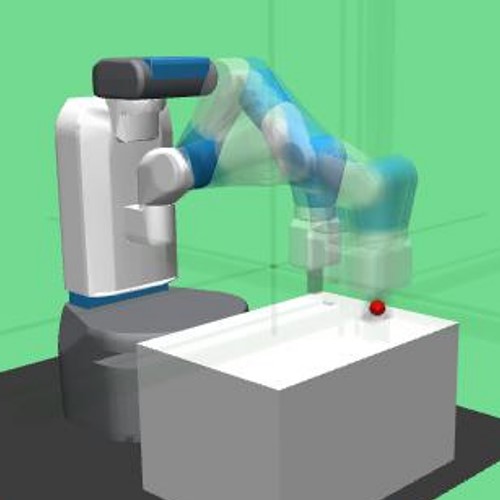}}
  \caption{Environments used in this work.}
  \label{fig:all_envs}
\end{figure}

\subsection{IBC Implementation}

\begin{table}[h]
\centering
\caption{Hyperparameters for IBC}
\label{table:hyperparameter_IBC}
\begin{tabular}{||c|c||c|c||}
\hline
critic hidden \textcolor{blue}{dimension} & 512 & discount factor $\gamma$ & 0.99 \\
critic hidden depth & 3 & curriculum buffer $\mathcal{B}_c$ \textcolor{blue}{capacity (\# of trajectories)} & 1000 \\
critic target $\tau$ & 0.01 & \# of curriculum candidates, K \textcolor{blue}{(\# of trajectories)} & 50 \\
critic target update frequency & 2 & curriculum update frequency \textcolor{blue}{(once every $N$ episode)} & 20 \\
actor hidden \textcolor{blue}{dimension} & 512 & learning rate & 1e-4 \\
actor hidden depth & 3 & RL optimizer & \textcolor{blue}{ADAM} \\
actor update frequency & 2 & init temperature $\alpha_\mathrm{init}$ of SAC & 0.5 \\
RL batch size & 512 & replay buffer $\mathcal{B}$ \textcolor{blue}{capacity (\# of transitions)}  & 1e6 \\
$c$ in curriculum update & 3 & Lipschitz constant $L$ & 5 \\
\hline
\end{tabular}
\end{table}

We use a goal-relabeling technique \cite{andrychowicz2017hindsight} with SAC for sparse reward, goal-conditioned RL. There is a separate trajectory-level buffer $\mathcal{B}_c$ for the bidirectional goal curriculum. The values of various hyperparameters are detailed in Table \ref{table:hyperparameter_IBC}. We set the goal space transformation of the auxiliary agent $\phi_a$ to abstract the proprioceptive states (e.g. gripper position in manipulation tasks and agent position in navigation or reaching tasks), and the goal space transformation of the forward agent $\phi_f$ to abstract the object-centric states when available (e.g. object position in manipulation tasks and agent position in navigation or reaching tasks).

\subsection{Baseline Implementations}

The baseline algorithms are trained as follows,

\begin{itemize}
    \item \textbf{VaPRL} \cite{sharma2021vaprl}: There is no official code implementation, so we implemented it ourselves. We closely followed the details in the original paper and validated whether we have properly implemented the algorithm by obtaining statistically similar results \textcolor{blue}{when using demonstrations} to the ones reported in \cite{sharma2021autonomous}.
    \item \textbf{MEDAL} \cite{sharma2022state}: We follow the default setting in the original implementation from \url{https://github.com/architsharma97/medal}. 
    \item \textbf{naive RL}: \textcolor{blue}{We train a single agent to reach the given goal state until success or} pre-determined, environment-specific maximum episode steps. After that, the target goal is resampled \textcolor{blue}{without resetting} and the agent repeats the above process for hundreds of thousands of steps. We use SAC \cite{haarnoja2018soft} with the goal relabeling technique \cite{andrychowicz2017hindsight}.
    \item \textbf{oracle RL}: Standard episodic RL is applied. Specifically, we use SAC \cite{haarnoja2018soft} with the goal relabeling technique \cite{andrychowicz2017hindsight}.
\end{itemize}

VaPRL and MEDAL require expert or near-expert demonstrations. For Sawyer Door and Tabletop Manipulation environments, we use the demonstration data (forward \& backward episodes) provided by the EARL benchmark \cite{sharma2021autonomous}. For other environments, we collected demonstrations of similar quality (expert-level) and quantity (comparable amount of total timesteps) by rolling out the trained oracle \textcolor{blue}{RL policy}.

\newpage
\section{Additional Experimental Results}\label{appendix:more_experiments}

\subsection{Sawyer Door with Velocity Inputs}\label{section:sawyer_door_vel}
For the Sawyer Door environment in the EARL benchmark \cite{sharma2021autonomous}, we found \textcolor{blue}{instances of the door moving} due to inertia even \textcolor{blue}{when the robot arm} is not in contact. \textcolor{blue}{Without velocity information, this} can violate the Markov Decision Process (MDP) assumption that the transition probability \textcolor{blue}{be} fully observable. That is, a different next state $s'$ can be obtained from the identical current state $s$ and action $a$. 

\begin{figure}[h]
    \centering
    \subfigure[Sawyer Door]{\includegraphics[width=0.31\columnwidth]{figures_matplotlib/0main_plot/sawyer_door.png}}
    \subfigure[Sawyer Door w Vel]{\includegraphics[width=0.31\columnwidth]{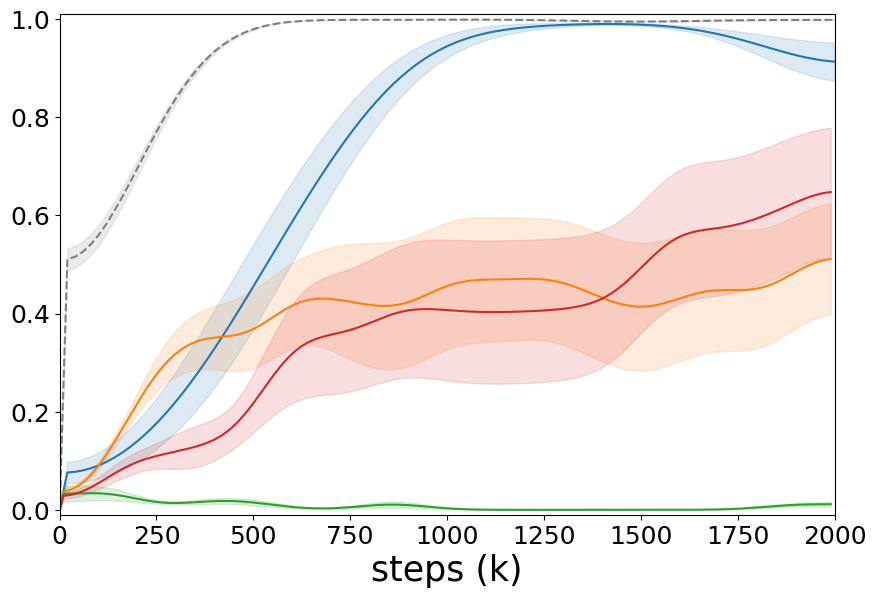}}
    \hfill
    \vskip -0.1cm
    \subfigure{\includegraphics[width=0.6\textwidth]{figures_misc/legend.png}}
    \vskip -0.3cm
  \caption{Experimental results for \textcolor{blue}{variants of} Sawyer Door environments.}
  \label{fig:sawyer_door_vel_results}
  \vskip -0.3cm
\end{figure}

To alleviate it, we additionally experiment with the velocity-augmented state inputs. We concatenate the translational velocity of the door handle (3-dimensional) and train the agent with IBC and other baselines. The results in Figure \ref{fig:sawyer_door_vel_results} demonstrate that there are slight increases in the final average success rates and training stability with velocity-augmented states.

\subsection{Sensitivity of Baseline Algorithms to the Demonstration Data}
\textcolor{blue}{Although} prior works (VaPRL, MEDAL) have shown some progress in developing a better ARL algorithm, these have some restrictions \textcolor{blue}{due to requiring} demonstration data for selecting the curriculum subgoals or \textcolor{blue}{for} computing the reward for the backward policy. Furthermore, we found that these baselines are somewhat sensitive to the composition of \textcolor{blue}{the} demonstration data.  

Specifically, we collected demonstration data for the Sawyer Door environment in two different ways. The first dataset consists of expert trajectories with fixed goal states (with trajectories of similar lengths), and the second dataset consists of expert trajectories with diverse goal states (with trajectories of varying lengths). Since we terminate the rollout right after the agent achieves the goal, trajectories may vary in length.

\begin{figure}[h]
    \centering
    \subfigure[MEDAL]{\includegraphics[width=0.31\columnwidth]{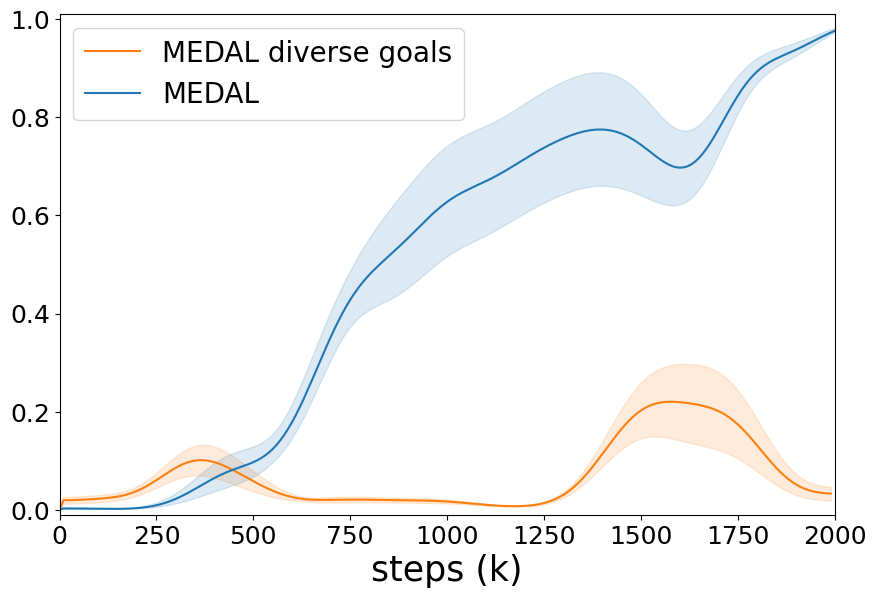}}
    \subfigure[VaPRL]{\includegraphics[width=0.31\columnwidth]{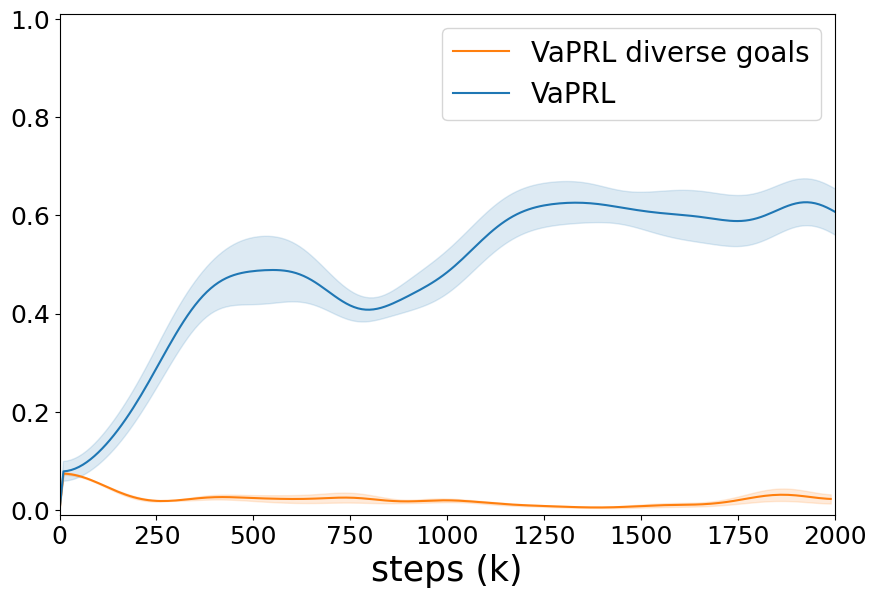}}
  \caption{Sensitivity of baselines to the composition of the demonstration data.}
  \label{fig:demo-sensitivity}
  \vskip -0.3cm
\end{figure}

As shown in Figure \ref{fig:demo-sensitivity}, both VaPRL and MEDAL are somewhat sensitive to the composition of the data. It may be due to the subgoal selection strategy in VaPRL that is dependent on the length of the demonstration trajectory, or in the case of MEDAL, an imbalance in the expert state distribution (for training the backward reward) caused by trajectories of varying lengths due to the diverse goals.

\vskip -0.3cm

\subsection{Episode Ratio of the Auxiliary Agent}

\begin{figure}[h]
    \centering
    
    \subfigure[Sawyer Door]{\includegraphics[width=0.245\columnwidth]{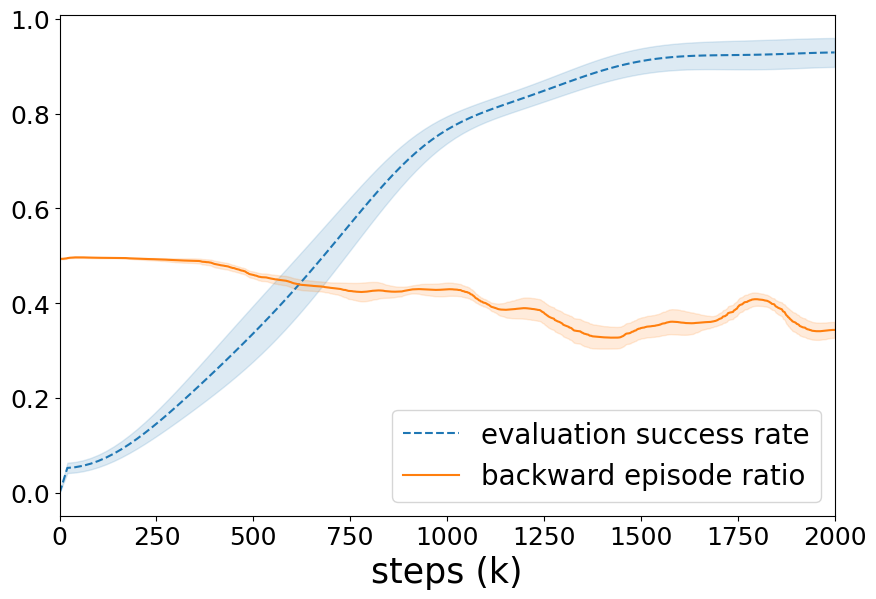}}
    \subfigure[Sawyer Door w Vel]{\includegraphics[width=0.245\columnwidth]{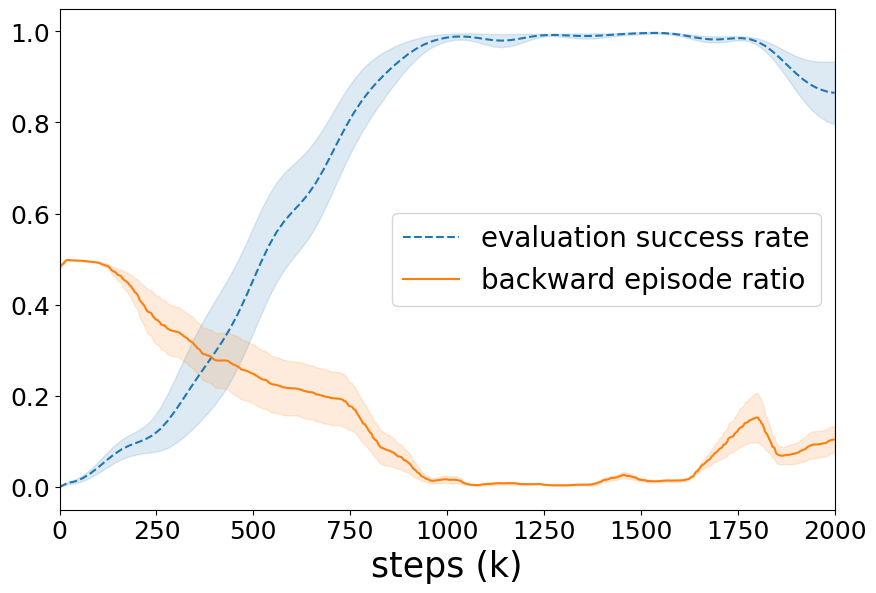}}
    \subfigure[Tabletop Manipulation]{\includegraphics[width=0.245\columnwidth]{figures_matplotlib/2fb_ratio_analysis_with_success_rate/tabletop_manipulation.png}}
    \subfigure[Point-U-Maze]{\includegraphics[width=0.245\columnwidth]{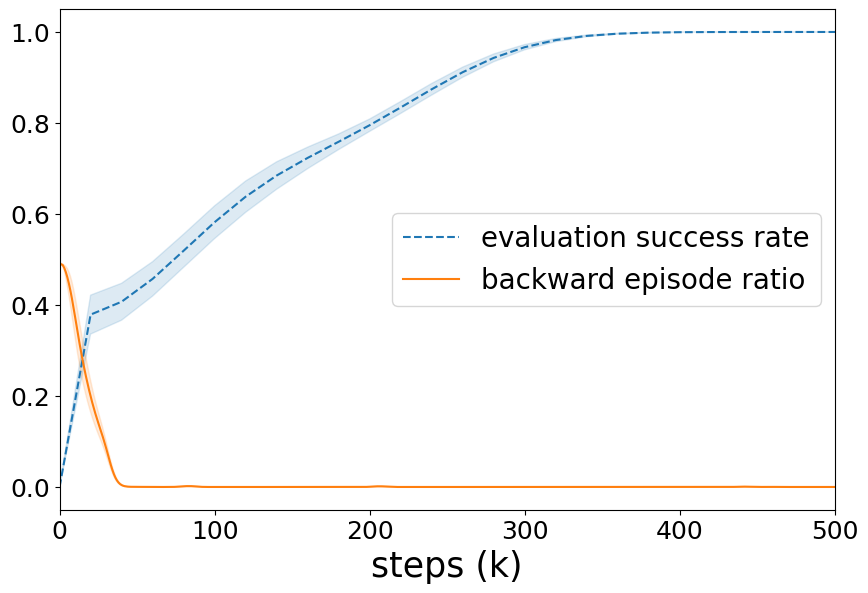}}
    
    \subfigure[Fetch Pick\&Place]{\includegraphics[width=0.245\columnwidth]{figures_matplotlib/2fb_ratio_analysis_with_success_rate/fetch_pickandplace_ergodic2.png}}
    \subfigure[Fetch Push]{\includegraphics[width=0.245\columnwidth]{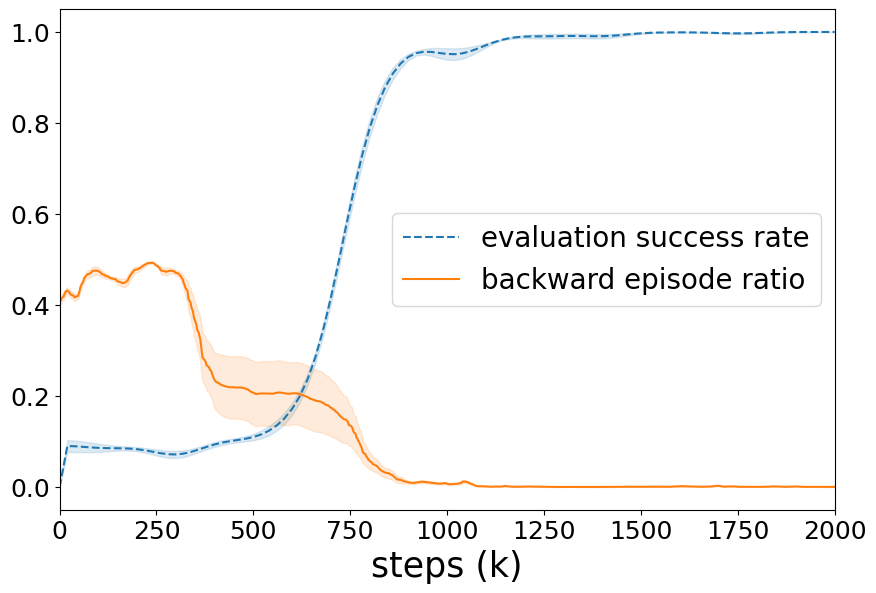}}
    \subfigure[Fetch Reach]{\includegraphics[width=0.245\columnwidth]{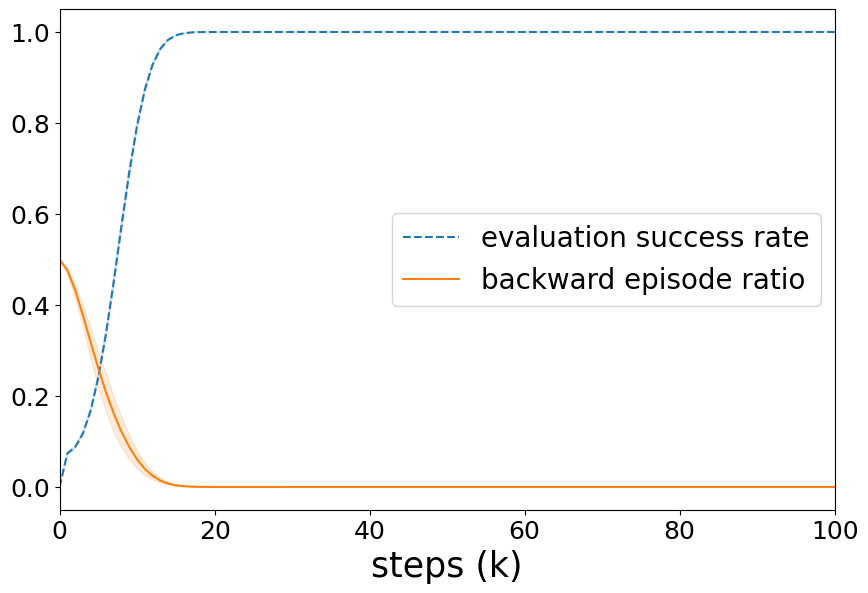}}

  \caption{Full results of auxiliary \textcolor{blue}{agent} episode ratio and evaluation success rate.}
  \label{fig:fb-ratio_full_results}
\end{figure}

We \textcolor{blue}{include} the full results of the auxiliary \textcolor{blue}{agent episode ratio} (Figure \ref{fig:fb-ratio_full_results}). The overall trend discussed in the main script also applies to the full result. We additionally report the results of Sawyer Door with velocity inputs as mentioned in \ref{section:sawyer_door_vel}, and we \textcolor{blue}{have found that the backward episode ratio decreases when} the velocity inputs are augmented.

\newpage

\subsection{Curriculum Visualization}

\begin{figure}[h]
    \centering
    \subfigure[Fetch Pick\&Place]{\includegraphics[width=0.48\textwidth]{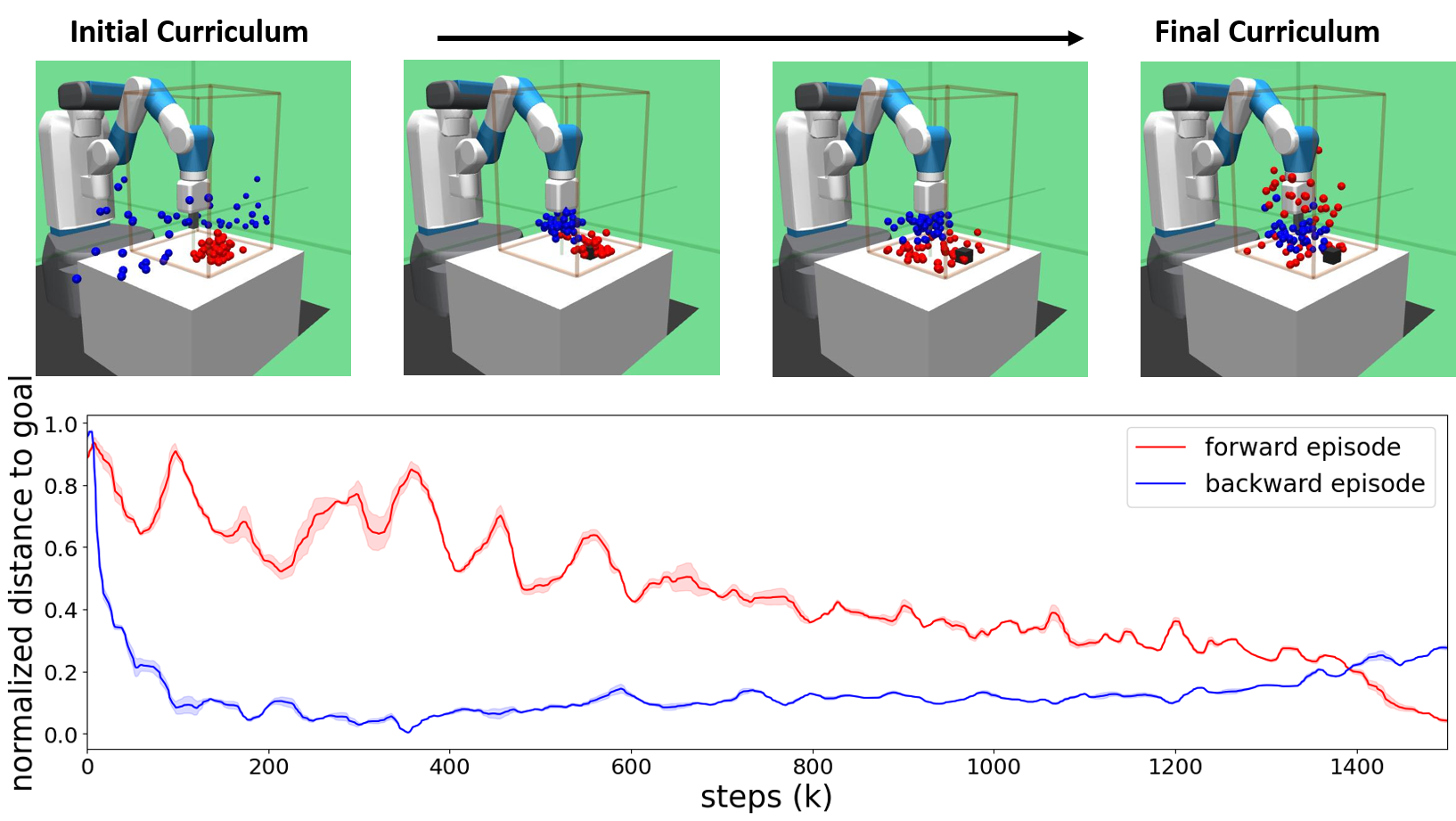}}
    \subfigure[Tabletop Manipulation]{\includegraphics[width=0.48\textwidth]{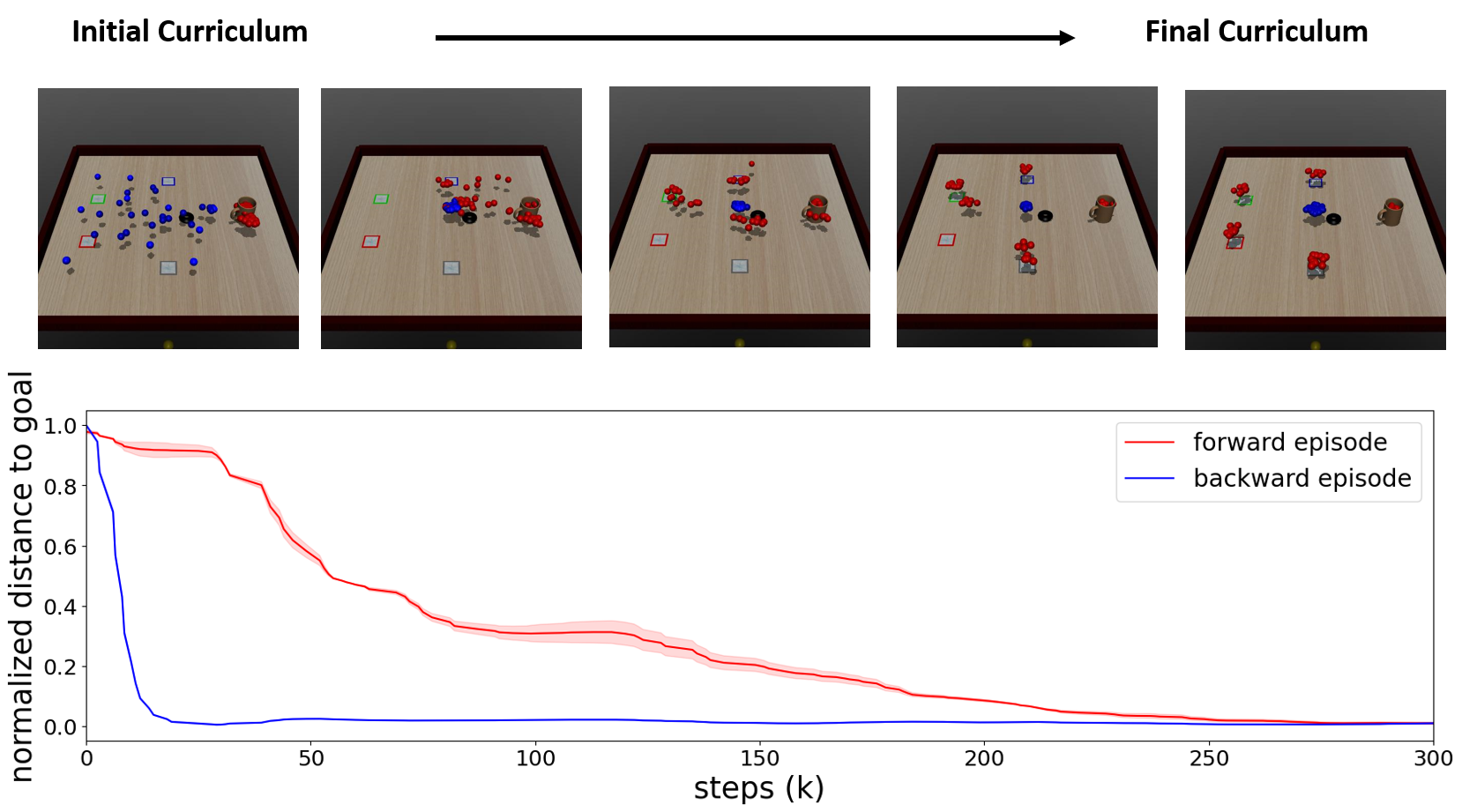}}

    \subfigure[Fetch Push]{\includegraphics[width=0.48\textwidth]{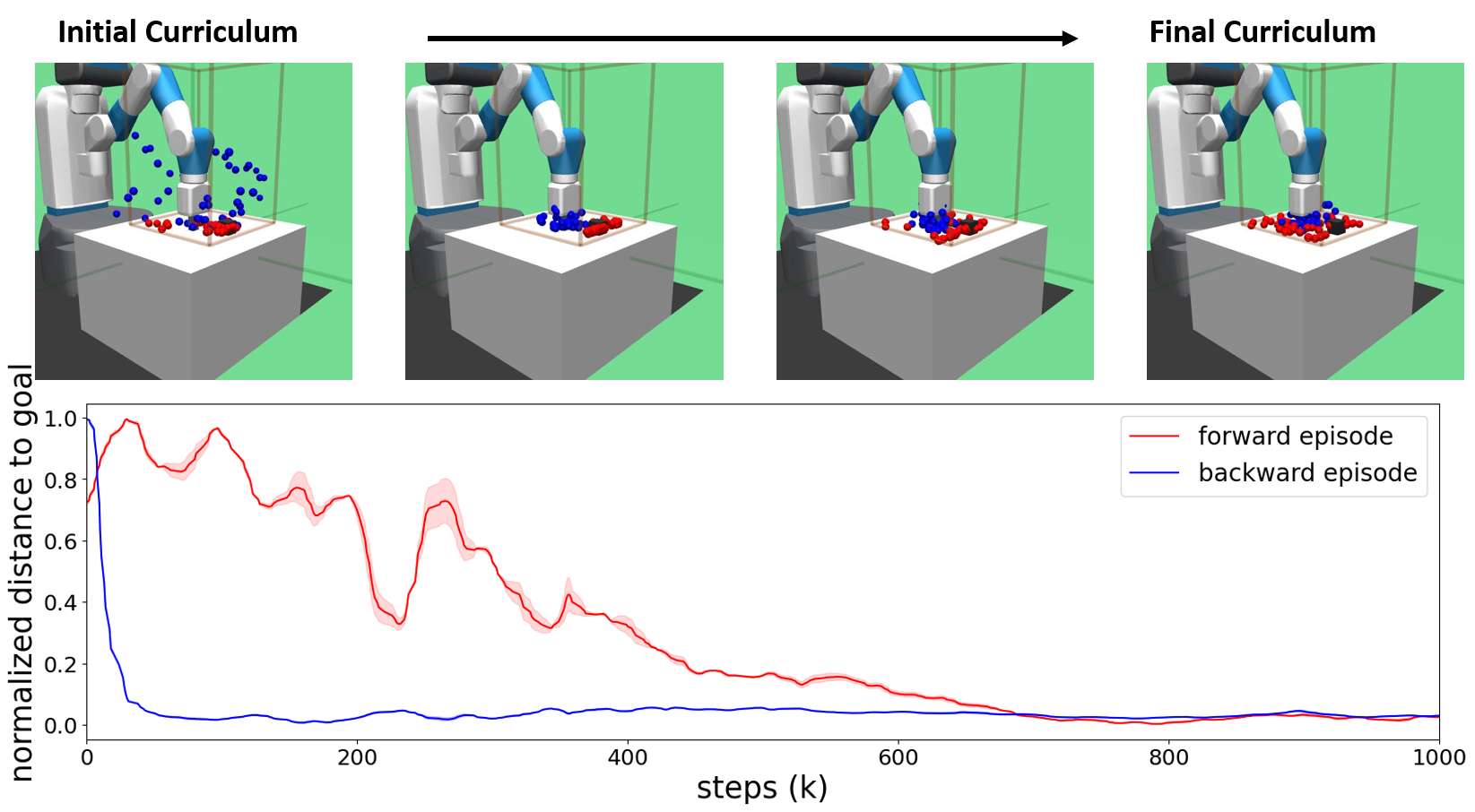}}
    \subfigure[Sawyer Door]{\includegraphics[width=0.48\textwidth]{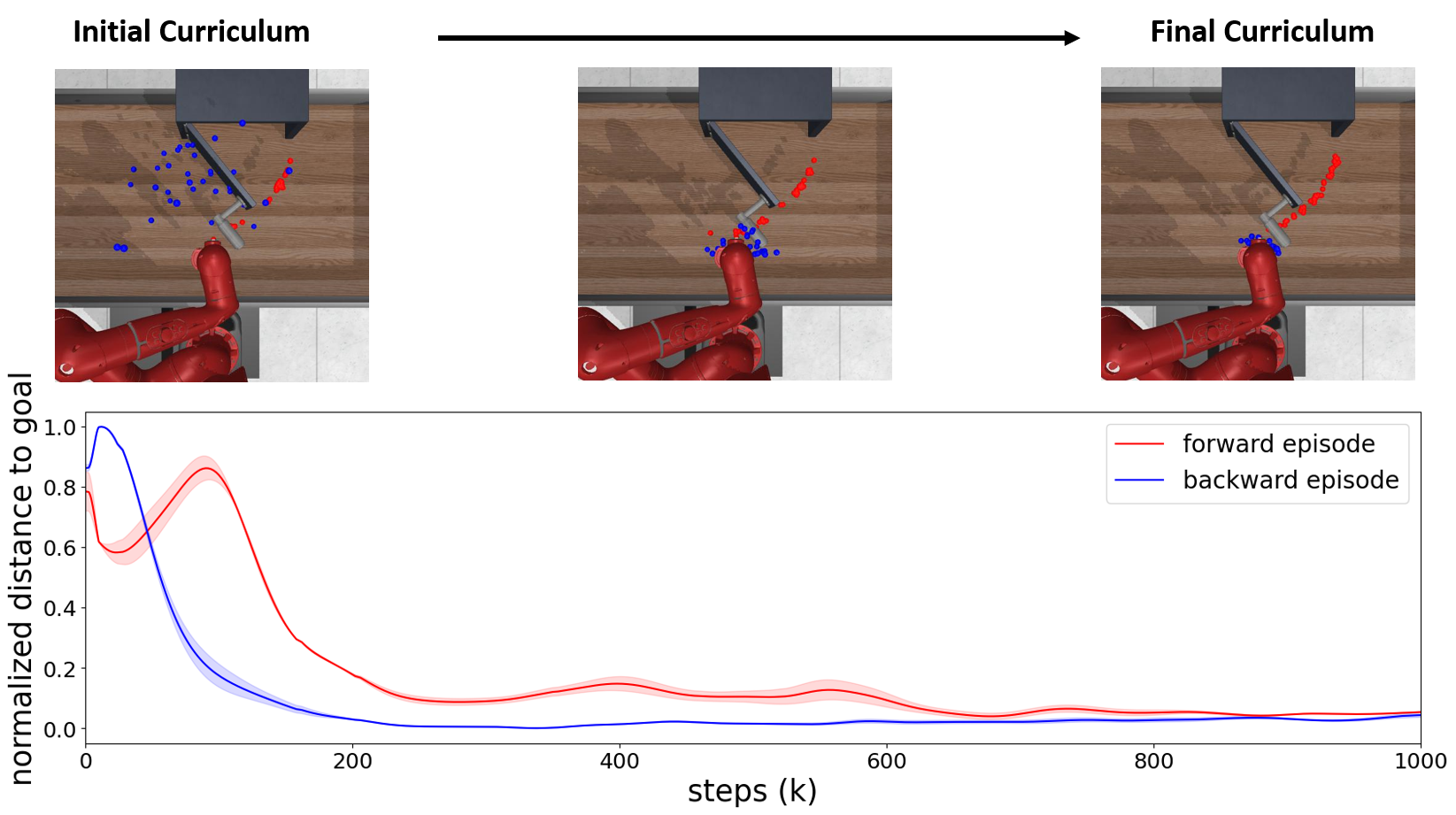}}

    \subfigure[Point-U-Maze]{\includegraphics[width=0.48\textwidth]{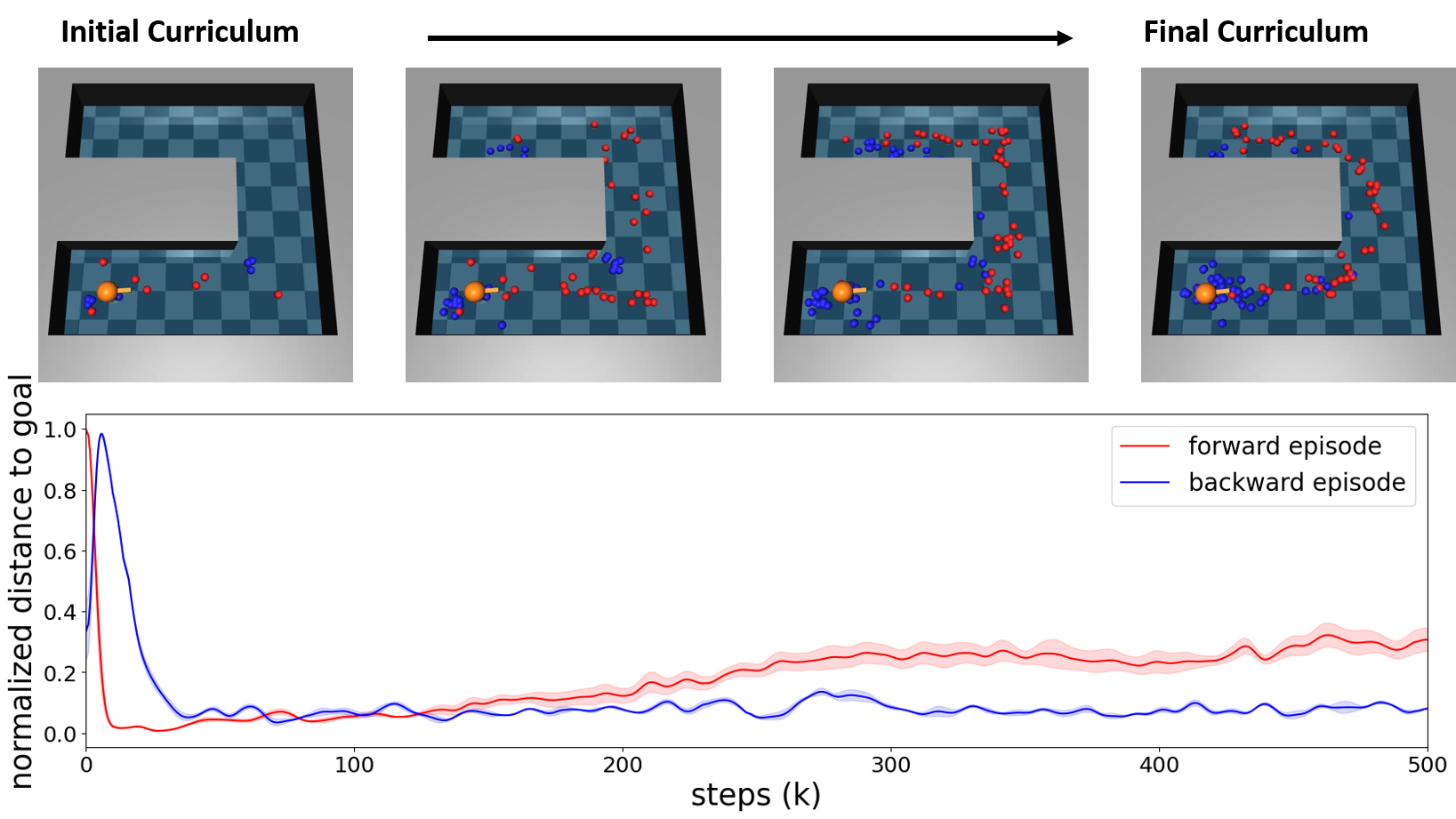}}

  \caption{Full results of the curriculum goals visualization and their average normalized distance to assigned target goals. The red and blue dots indicate the curriculum goals for the forward and auxiliary agents, respectively. Note that the exact positions of the robots and objects are meaningless; these are just rendered from their default states.}
  \label{fig:curriculum-visualize_full_results}
\end{figure}

We report the full results for the curriculum visualization from the main \textcolor{blue}{script. As shown in Figure \ref{fig:curriculum-visualize_full_results}, curriculum goals} for both forward and auxiliary agents \textcolor{blue}{converge} to their respective target state distribution.

\newpage
\subsection{Ablation Study}

\begin{figure}[h]
    \centering

    \vskip -0.25cm
    \subfigure[Sawyer Door]{\includegraphics[width=0.245\textwidth]{figures_matplotlib/1ablation/sawyer_door.png}}
    \subfigure[Sawyer Door w Vel]{\includegraphics[width=0.245\textwidth]{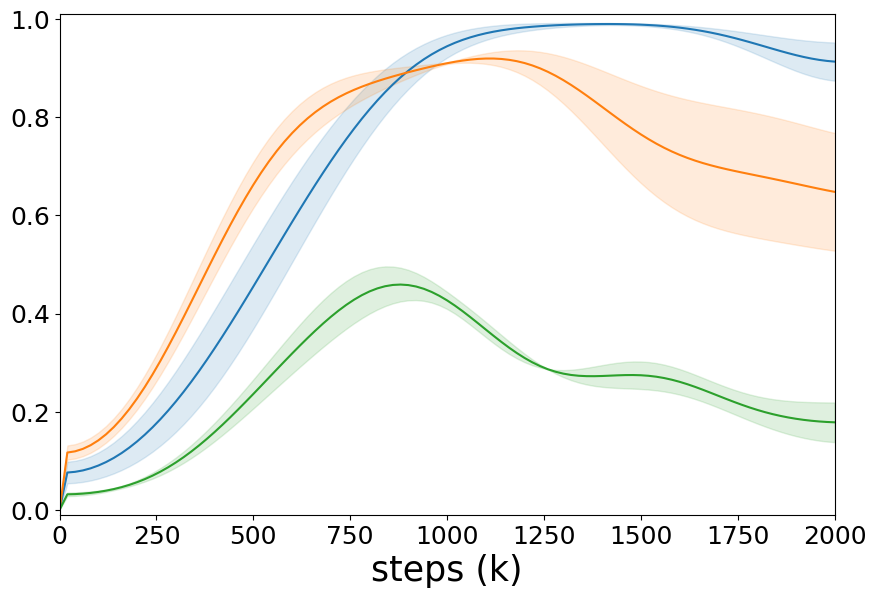}}
    \subfigure[Tabletop Manipulation]{\includegraphics[width=0.245\textwidth]{figures_matplotlib/1ablation/tabletop_manipulation.png}}
    \subfigure[Point-U-Maze]{\includegraphics[width=0.245\textwidth]{figures_matplotlib/1ablation/point_umaze.png}}
    \subfigure[Fetch Pick\&Place]{\includegraphics[width=0.245\textwidth]{figures_matplotlib/1ablation/fetch_pickandplace_ergodic2.png}}    
    \subfigure[Fetch Push]{\includegraphics[width=0.245\textwidth]{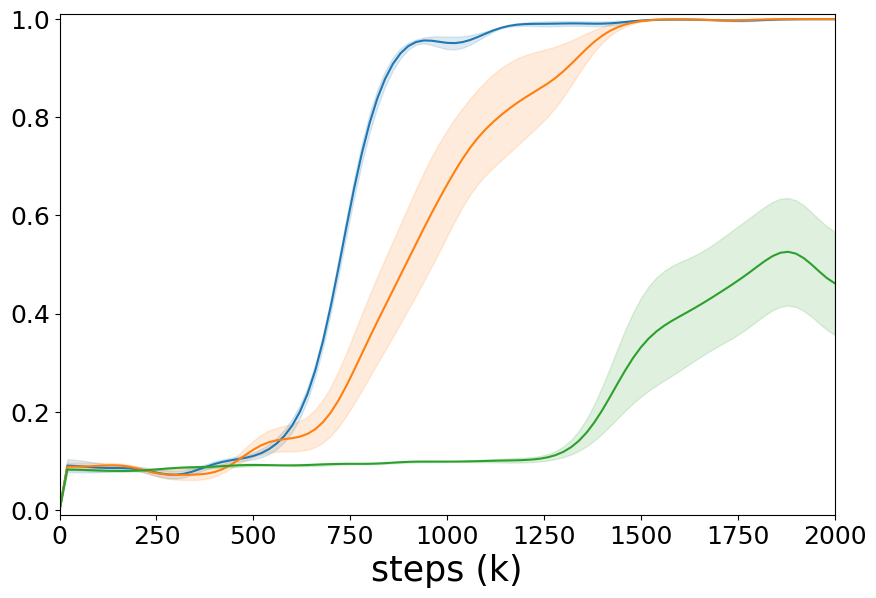}}
    \subfigure[Fetch Reach]{\includegraphics[width=0.245\textwidth]{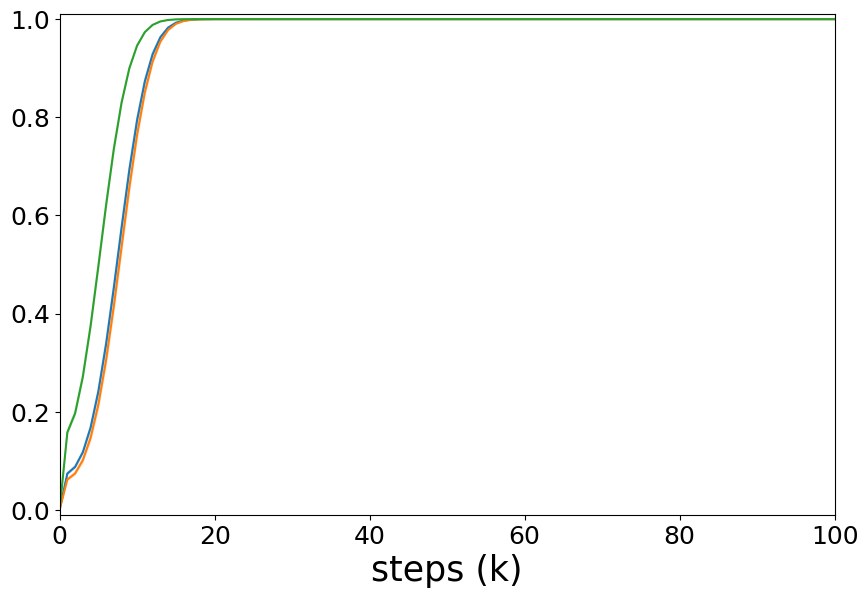}}    
    
    \subfigure{\includegraphics[width=0.5\textwidth]{figures_misc/legend_ablation.png}}
  \caption{Full results of ablation study -- removing the bidirectional curriculum and auxiliary agent proposed in this work degrades performance.}
  \label{fig:ablation_full_results}
\end{figure}

We report the full results of the ablation study from the main script. As shown in Figure \ref{fig:ablation_full_results}, success rates generally decrease as we remove the bidirectional goal curriculum, with further degradation when the auxiliary agent is removed.

\subsection{Reward-free Variant}

\begin{figure}[h]
    \centering

    \vskip -0.25cm  
    \subfigure[Sawyer Door]{\includegraphics[width=0.245\columnwidth]{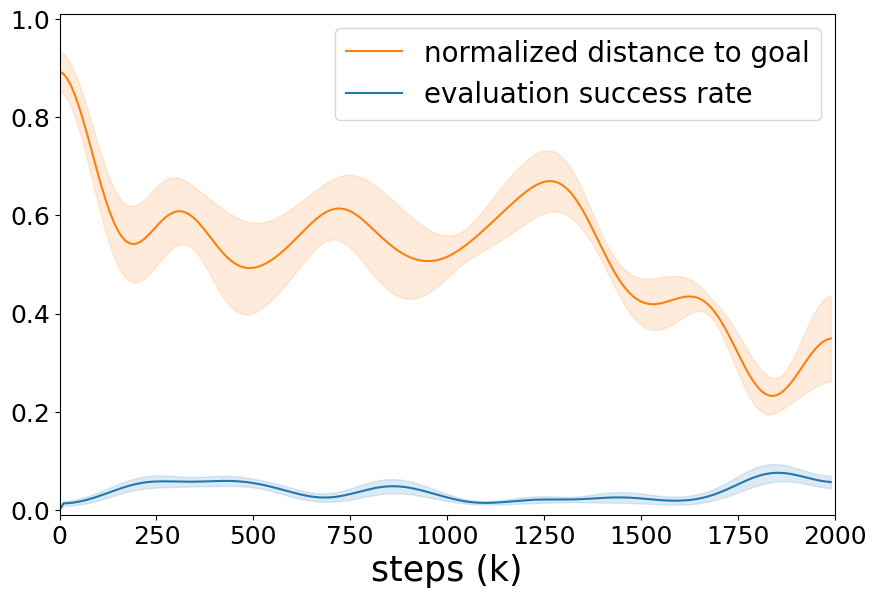}}
    \subfigure[Sawyer Door w Vel]{\includegraphics[width=0.245\columnwidth]{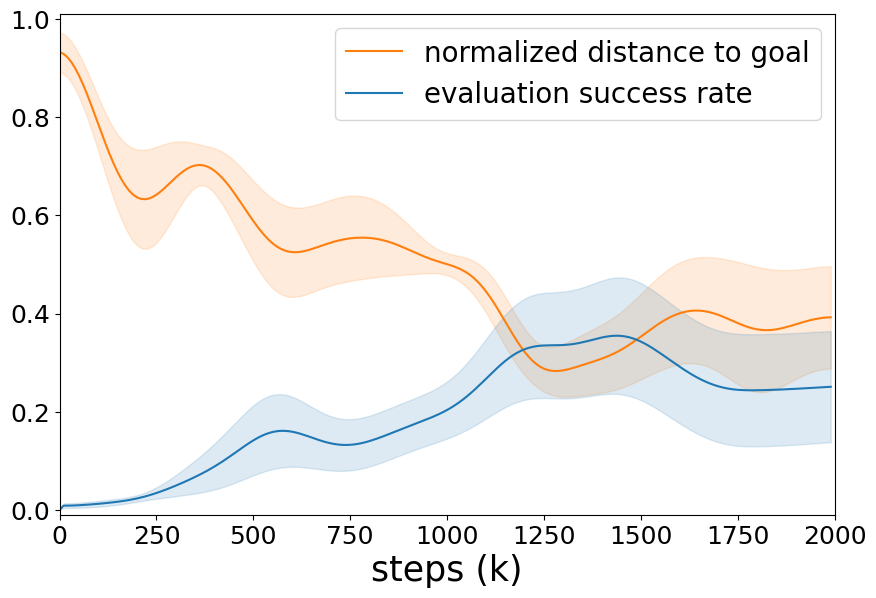}}
    \subfigure[Point-U-Maze]{\includegraphics[width=0.245\columnwidth]{figures_matplotlib/2reward_free_analysis/point_umaze.png}}
    \subfigure[Tabletop Manipulation]{\includegraphics[width=0.245\columnwidth]{figures_matplotlib/2reward_free_analysis/tabletop_manipulation.png}}
    \vskip -0.01cm
    \subfigure[Fetch Pick\&Place]{\includegraphics[width=0.245\columnwidth]{figures_matplotlib/2reward_free_analysis/fetch_pickandplace_ergodic2.png}}
    \subfigure[Fetch Push]{\includegraphics[width=0.245\columnwidth]{figures_matplotlib/2reward_free_analysis/fetch_push_ergodic2.png}}
    \subfigure[Fetch Reach]{\includegraphics[width=0.245\columnwidth]{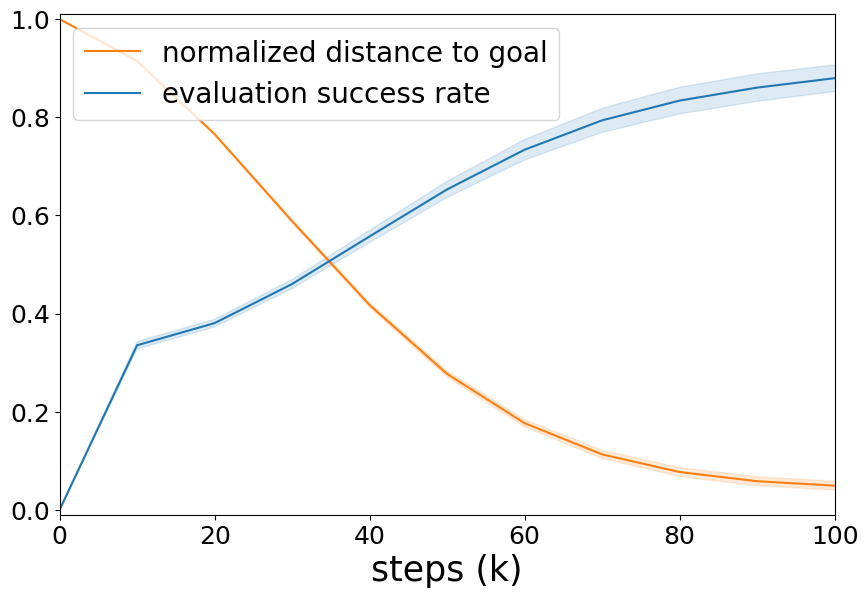}}
  \caption{Full results of the reward-free variant -- normalized distance to goal and evaluation success rate.}
  \label{fig:reward-free_full_results}
\end{figure}

We report the full results of the reward-free variant (C-learning variant) from the main script in Figure \ref{fig:reward-free_full_results}. The overall trend discussed in the main script also applies to the full result. One thing of note is that the reward-free variant performs better in the Sawyer Door environment when the door velocity is \textcolor{blue}{augmented}.

\end{document}